\pdfoutput=1
\documentclass[runningheads]{llncs}

\usepackage[final]{eccv}

\usepackage{eccvabbrv}
\usepackage{csquotes}
\usepackage{graphicx}
\usepackage{booktabs}
\usepackage{multirow} 
\usepackage{wrapfig}
\usepackage{makecell}
\usepackage{colortbl}
\usepackage{marvosym}
\usepackage[accsupp]{axessibility}  

\definecolor{rankone}{HTML}{DDF1D4}
\definecolor{ranktwo}{HTML}{FFFAD4}
\definecolor{rankthree}{HTML}{C2DEEB}
\usepackage{hyperref}
\hypersetup{
	pdftitle={AnyMatch: Supercharging Universal Multi-Modal Image Matching with Large-Scale Single-View Images},
	pdfauthor={Meng Yang, Zizhuo Li, Linfeng Tang, Fan Fan, Jiayi Ma}
}

\usepackage{orcidlink}
\graphicspath{{figure/}{supp/figure/}}

\begin{document}

\title{AnyMatch: Supercharging Universal Multi-Modal Image Matching with Large-Scale Single-View Images} 
\titlerunning{AnyMatch}

\author{Meng Yang\inst{1}$^*$    
\orcidlink{0000-0001-6382-2181}   \and
Zizhuo Li\inst{1}$^*$$^\dag$ \orcidlink{0000-0003-0986-4924} \and
Linfeng Tang\inst{2}\orcidlink{0000-0002-8566-5743} \and
Fan Fan\inst{2} \textsuperscript{\Letter}  \orcidlink{0000-0002-7507-1810} \and
Jiayi Ma\inst{2}\orcidlink{0000-0003-3264-3265} 
}

\authorrunning{M.~Yang et al.}

\institute{Electronic Information School\inst{1}, Wuhan University, Wuhan 430072, China\\
	School of Robotics, Wuhan University\inst{2}, Wuhan 430072, China\\
\email{\{2024102120059, zizhuo\_li, fanfan\}@whu.edu.cn, \{linfeng0419, jyma2010\}@gmail.com}}
\footnotetext{Equal contribution. $^\dag$ Project leader. \textsuperscript{\Letter} Corresponding author.}
\maketitle

\begin{abstract}
Multi-modal image matching is essential for visual localization and multi-sensor fusion, but it is hindered by the scarcity of large-scale training data with precise geometric annotations. Existing real-world datasets suffer from prohibitive costs, limited scene diversity, and errors in SfM-MVS pipelines, while synthetic methods struggle to maintain 3D geometric consistency or achieve photorealistic appearance. To address this, we propose AnyMatch, a novel framework that leverages abundant, easily accessible single-view images at minimal cost to generate rich multi-modal training data. AnyMatch integrates monocular depth estimation, 3D reprojection, diffusion-based inpainting, and cross-modal image translation to synthesize multi-view, multi-modal image pairs with 3D geometric fidelity. Crucially, our method provides annotations that strictly adhere to 3D geometric consistency through explicit 3D reprojection, avoiding SfM-MVS error accumulation. Furthermore, AnyMatch offers strong scalability, enabling controllable scene diversity and annotation difficulty via adjustable input and camera parameters. We construct Any-syn, a large-scale synthetic multi-modal dataset using AnyMatch. Experimental results show that matching networks (\textit{e.g.}, LoFTR, EDM, RoMa) fine-tuned on Any-syn achieve substantial performance gains on multi-modal benchmarks, exhibiting superior generalization and robustness compared to models trained on existing data.

\end{abstract}

\section{Introduction}
\label{sec:intro}

Feature matching across different imaging modalities is a cornerstone of many computer vision systems, enabling critical capabilities such as visual localization under varying conditions \cite{zhou2021vmloc}, cross-spectral object detection \cite{liu2022target}, and multi-sensor fusion for autonomous systems \cite{aditya2024thermal}. Traditional feature matching methods, from handcrafted descriptors like SIFT \cite{lowe2004distinctive} and SURF \cite{bay2006surf} to modern deep learning approaches like LoFTR \cite{sun2021loftr}, EDM \cite{li2025edm} and RoMa \cite{edstedt2024roma}, have primarily focused on single-modality (RGB-RGB) scenarios. However, real-world applications often require matching between images captured by different sensors (\textit{e.g.}, RGB-thermal, RGB-depth, or RGB-event cameras), where modal differences dramatically challenge conventional single-modal matching methods.
\begin{figure*}[t]	
	\centering	
	\includegraphics[scale=0.132]{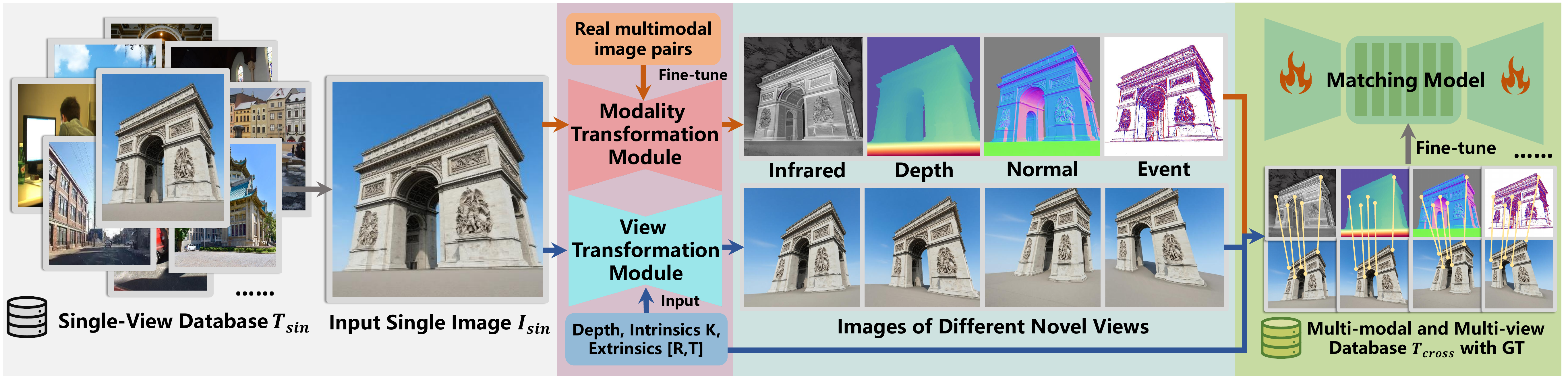}
	\caption{Overview of AnyMatch pipeline. AnyMatch aims to generate a large-scale multi-modal and multi-view image matching dataset for training matching models, thereby enabling robust cross-modal and cross-view matching capabilities.}\label{fig1}
\end{figure*}

The fundamental bottleneck hindering the development of robust multi-modal matching algorithms lies in the scarcity of large-scale training data with precise geometric annotations. However, both existing real-world and synthetic multi-modal datasets suffer from significant limitations.

Existing multi-modal image datasets face three critical challenges: prohibitive acquisition costs, limited scene diversity, and inaccurate geometric annotations. First, capturing genuine multi-modal image pairs requires specialized, synchronized sensor suites (\textit{e.g.}, RGB and thermal cameras) with precise cross-modal calibration, resulting in substantial hardware and labor expenditures. Second, geometric ground truth (GT) for existing single-modal datasets is typically derived via Structure-from-Motion (SfM) and Multi-View Stereo (MVS) pipelines from multi-view imagery \cite{li2018megadepth,dai2017scannet}. However, such reconstruction approaches inherently favor static scenes with high appearance consistency, minimal occlusion, and good illumination conditions, thereby failing to reflect the complexity of real-world environments and limiting coverage of diverse scene content. These constraints also preclude direct application of SfM-MVS pipelines to GT generation of multi-modal images. Third, SfM-MVS-derived datasets suffer from error accumulation—including pose estimation inaccuracies, reconstruction failures in textureless regions, and violations of photometric consistency assumptions—yielding sparse and unreliable geometric constraints and depth estimation. Consequently, models trained on such data, which exhibit severely limited diversity and accuracy, often undergo catastrophic performance degradation when deployed in unseen scenarios or under challenging conditions.

To mitigate the scarcity of real-world data, existing synthetic data generation methods offer a viable alternative. First, image translation–based synthesis methods such as MINIMA \cite{ren2025minima} and MatchAnything \cite{he2025matchanything} synthesize multi-modal image pairs by applying cross-modal translation to pre-existing single-modal matching datasets (\textit{e.g.}, MegaDepth \cite{li2018megadepth}), followed by fine-tuning to improve matching performance on the generated data. 
However, it relies on SfM for dataset construction, with three limitations: static-region reconstruction only (scene biases), tourist attractions-centric scenes (limited diversity), and time-consuming multi-view image acquisition. Second, 2D homography–based synthesis methods operate solely within the image plane via homography transformations \cite{deng2024crosshomo,zhang2024sparse}, thereby discarding cross-view 3D geometric consistency (such as the physical relationship between parallax and depth) and failing to produce occlusion patterns that adhere to real-world physics. Third, game engine–driven synthesis methods \cite{mayer2016large,greff2022kubric} enable explicit control over scene parameters but struggle to achieve photorealistic appearance and accurate physical effects, particularly in material reflectance properties and global illumination. Although synthetic data generation circumvents the high costs of real data acquisition and manual annotation, thereby significantly reducing dataset construction expenses, it introduces non-negligible limitations. These constraints restrict the scale, geometric accuracy, and physical authenticity of current synthetic multi-modal datasets.

In contrast to these limitations, vast repositories of single-view images are readily available on the internet and across diverse research domains. Datasets such as GLDv2 \cite{weyand2020google}, SA-1B \cite{kirillov2023segment}, and COCO \cite{lin2014microsoft} collectively comprise millions of images spanning heterogeneous environments—including indoor/outdoor settings and urban/natural scenes—under varying conditions and domains. These resources inherently embody the complexity and long-tail distribution characteristics of the real world, theoretically offering feature matching models unprecedented scene coverage and generalization potential. Nevertheless, the absence of an effective mechanism to synthesize multi-view, multi-modal image pairs with 3D spatially consistent geometry and pixel-level precise supervision has left these abundant resources in a state of untapped potential: despite their richness, they remain largely inaccessible for advancing visual matching tasks.

Inspired by recent advances in monocular depth estimation, image generation, and cross-modal image translation, we propose AnyMatch, a novel framework that transforms abundant, easily accessible single-view images at minimal cost into a rich source of multi-modal training data with high 3D geometric fidelity. AnyMatch comprises two core components: view transformation and modality transformation, as shown in Fig.~\ref{fig1}. The view transformation module lifts a single 2D image into 3D space by estimating depth and camera intrinsics, followed by novel view synthesis via 3D reprojection based on the camera extrinsics, as shown in Fig.~\ref{fig2}. Using diffusion-based inpainting, we complete occluded regions to produce visually coherent novel views. The modality transformation module generates different modal images (infrared, depth, normal and event modalities) through cross-modal image translation techniques. By jointly performing view transformation and modality transformation, we synthesize multi-modal image pairs and annotations that strictly adhere to 3D geometric consistency, avoiding SfM-MVS error accumulation. This pipeline can be applied to any existing image dataset, enabling the generation of arbitrarily large, diverse multi-modal training sets with pixel-perfect geometric supervision.

AnyMatch offers three distinct advantages over existing methods. First, it decouples data generation from expensive multi-sensor hardware by leveraging the abundance of readily accessible single-view images at minimal cost. Second, it provides stereo-geometrically precise GT through explicit 3D reprojection, ensuring annotations that strictly adhere to 3D geometric consistency and avoiding error accumulation. Third, it can control scene diversity and annotation difficulty levels via adjustable input and camera parameters, thereby supporting diverse application requirements and facilitating curriculum learning strategies.

To validate our method, we fine-tune existing image matching networks (LoFTR, EDM, and RoMa) on datasets synthesized by AnyMatch. Experimental results demonstrate that the fine-tuned models achieve substantial performance gains on cross-modal benchmark evaluations.

In summary, our contributions are threefold:
\begin{itemize}
\item We propose AnyMatch, the first framework capable of synthesizing large-scale multi-modal image pairs with high 3D geometric fidelity from abundant, easily accessible single-view images at minimal cost.

\item We present a novel pipeline that integrates monocular depth estimation, 3D reprojection, diffusion-based inpainting, and cross-modal image translation to synthesize diverse multi-modal datasets with annotations that strictly adhere to 3D geometric consistency.

\item We construct Any-syn, a synthetic multi-modal dataset generated via AnyMatch. Matching networks fine-tuned on Any-syn achieve superior performance on multi-modal matching benchmarks, with significant improvements in generalization and robustness. The supplementary material presents some data synthesis results.
\end{itemize}
\section{Related Work}
\label{sec:Related}
\subsection{Multi-modal Image Matching}
Cross-modal image matching aims to establish pixel-wise correspondences between images captured under different sensing modalities, serving as a fundamental building block for visual localization \cite{zhou2021vmloc,aditya2024thermal} and multi-sensor fusion tasks \cite{xu2023murf,zhang2021image}. Traditional methods primarily rely on handcrafted feature descriptors that extract shape-, gradient-, or phase-based features \cite{hou2024pos,li2019rift,gao2025homo,li2023multimodal}, yet exhibit limited robustness in textureless regions or under drastic illumination variations. Recently, data-driven deep learning methods have demonstrated superior feature representation capabilities. Existing works typically adopt off-the-shelf matching architectures (\textit{e.g.}, EDM \cite{li2025edm}, LoFTR \cite{sun2021loftr}, and RoMa \cite{edstedt2024roma}) as backbone networks, with task-specific adaptations for target modalities. For instance, ReDFeat \cite{deng2022redfeat} decouples detection and description constraints in multi-modal feature learning via a mutual weighting strategy. XoFTR \cite{tuzcuouglu2024xoftr} and XCP-Net \cite{yang2025multimodal} employ a two-stage training paradigm that first pre-trains on real multi-modal data and subsequently fine-tunes on pseudo-infrared images, significantly boosting RGB-infrared matching performance. SCFeat \cite{cadar2024leveraging} and SAMFeat \cite{wu2025segment} leverage semantic priors extracted from vision foundation models (\textit{e.g.}, DINOv2 \cite{oquab2023dinov2} and SAM \cite{kirillov2023segment}) to refine feature representation and enhance matching accuracy. PYM translates cross-modal images into the matcher’s native domain via correspondence-aware adversarial learning, enabling competitive cross-modal matching without modifying the pre-trained matcher \cite{frolov2026patch}. Nevertheless, these methods are predominantly tailored to specific modality pairs and remain severely constrained by the scale and diversity of available training data, hindering their generalization capability across unseen modalities and real-world scenarios.
\subsection{Synthetic Data Generation for Matching}
Current large-scale RGB image matching datasets (\textit{e.g.}, MegaDepth~\cite{li2018megadepth}, ScanNet~\cite{dai2017scannet}) primarily rely on SfM-MVS pipelines to generate geometric GT. MegaDe\-pth reconstructs scene depth and camera poses from internet-sourced multi-view photographs using open-source tools such as COLMAP \cite{schonberger2016structure,schonberger2016pixelwise}. ScanNet similarly generates annotations from indoor RGB-D video sequences through analogous reconstruction workflows. The core advantage of these methods lies in their ability to leverage real-world imagery and mature geometric reconstruction tools to derive correspondence supervision. However, the SfM-MVS pipeline suffers from several inherent limitations: (1) it intrinsically favors static scenes with high photometric consistency, minimal occlusions, and favorable lighting conditions, struggling with dynamic objects, drastic illumination changes, or textureless regions; (2) the multi-stage reconstruction process accumulates errors across stages; and (3) it critically depends on multi-view image inputs. Owing to significant discrepancies in pixel intensity distributions and fundamental gaps in imaging mechanisms across modalities, conventional SfM-MVS pipelines fail to operate robustly on multi-modal image pairs. Consequently, acquiring accurate geometric annotations for real-world multi-modal data remains highly challenging. Coupled with the high cost of acquiring real multi-modal data, this has led to a severe scarcity of annotated multi-modal image matching datasets. To mitigate this data scarcity, researchers have proposed diverse synthetic data generation strategies, which can be broadly categorized into three classes:
\subsubsection{Image translation-based synthesis methods.}
These methods generate multi-modal image pairs by applying cross-modal translation to existing single-modality matching datasets. As a representative work, MINIMA \cite{ren2025minima} proposes a data synthesis engine that leverages existing multi-view RGB datasets and employs generative models (StyleBooth \cite{han2025stylebooth} and DepthAnything v2 \cite{yang2024depth}) to synthesize pseudo-modality images including infrared (IR), depth, and event modalities, \textit{etc}. The geometric GT (depth maps and camera poses) from the source dataset is directly inherited to serve as correspondence supervision for the synthesized multi-modal pairs. This method demonstrates that leveraging large-scale synthetic multi-modal data can effectively enhance the cross-modal generalization capability of matching models. Nevertheless, MINIMA exhibits inherent limitations: (1) its source data is constrained by the scene coverage and scale of existing multi-view datasets; (2) its geometric ground truth is directly inherited from SfM-MVS reconstruction results, inevitably subject to pose estimation errors and reconstruction failures. These constraints propagate into the synthesized multi-modal pairs, compromising the geometric accuracy of the training supervision.
\subsubsection{Homography-based synthesis methods.}
These methods simulate geometric deformations by applying homography transformations within the image plane \cite{deng2024crosshomo,zhang2024sparse}. Although computationally efficient and straightforward to implement, they discard cross-view 3D geometric consistency and fail to generate occlusion patterns that conform to real-world physics. As a result, models trained on such synthetic data exhibit weak generalization capability under view variations encountered in real-world scenarios.
\subsubsection{Game engine-based synthesis methods.} 
Several works leverage game engines to generate synthetic data \cite{mayer2016large,greff2022kubric}. While these methods enable explicit control over scene parameters and sensor models, they struggle to achieve photorealistic appearance fidelity—particularly in modeling material reflectance properties, global illumination effects, and sensor noise characteristics. This results in a significant domain gap between synthesized and real-world images, thereby limiting the transferability of models trained on such data to practical applications.

To address the aforementioned limitations, we propose AnyMatch, a framework that synthesizes multi-view, multi-modal image pairs with high 3D geometric fidelity from large-scale real-world single-view images. Compared with the existing data synthesis methods, AnyMatch utilizes real scene images, does not rely on expensive acquisition hardware, and provides GT that strictly adheres to 3D geometric consistency, without the error accumulation of SfM-MVS.

\section{Method}
\label{sec:method}
Existing image matching datasets predominantly rely on real-world multi-view captures; however, they are inherently constrained by limited data scale, insufficient scene diversity, narrow modality coverage, and sub-optimal annotation quality. These limitations hinder the development and training of robust, general-purpose multi-modal matching models. To address this challenge, we propose AnyMatch, a geometry-supervised generative framework for multi-modal image synthesis. The framework explicitly integrates two core components: view transformation and modality transformation.
\subsection{View Transformation}
We construct a large-scale, diverse, and high-quality single-view image dataset from publicly available internet sources across multiple research domains (\textit{e.g.}, GLDv2 and SA-1B), predominantly in the visible modality, to establish the single-view database \( T_{\text{sin}} \). All images are uniformly resized to $512\times 512$ via isotropic scaling and center cropping. As shown in Fig.~\ref{fig2}, AnyMatch takes a single-view image \( I_{\text{sin}} \in T_{\text{sin}} \) as input. First, we apply a pre-trained monocular relative depth estimation model \( M_{\text{rd}} \) to predict the dense relative depth map \( D_r = M_{\text{rd}}(I_{\text{sin}}) \), which captures the depth of fine-grained geometric structures of the image and preserves rich local details. Due to the inherent scale ambiguity in monocular depth estimation, we apply random affine perturbations to the depth to enhance the algorithm's robustness against depth scale uncertainty. Specifically, the normalized depth map $D_r \in [0,1]$ is transformed via:
\begin{figure*}[t]	
	\centering	
	\includegraphics[scale=0.134]{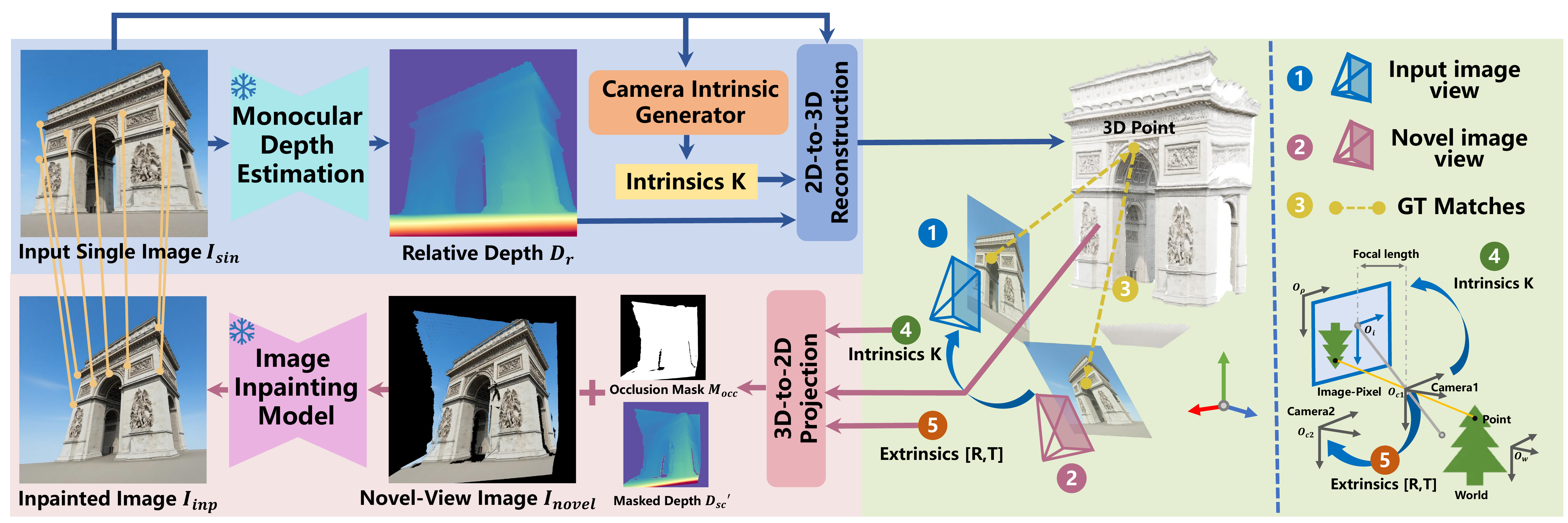}
	\caption{Overview of view transformation module. The view transformation module is designed to generate an arbitrary number of multi-view image pairs that strictly adhere to 3D geometric consistency.}\label{fig2}
\end{figure*}
\begin{equation}\label{equal1}
 D = 1/({\alpha \cdot(1/D_r) + \beta + \epsilon}), 
\end{equation}
where the scaling factor $\alpha \in (0.5, 2.0)$ simulates scale uncertainty arising from variations in camera position and focal length. The bias $\beta \in (0, 0.3)$ models sensor systematic errors and the non-zero mean noise inherent in the depth estimation network. $\epsilon$ is introduced for numerical stability. Subsequently, to lift the 2D image into 3D space, we design a random camera intrinsic generator. Specifically, we initialize a intrinsic ${K} = \left[\begin{smallmatrix} f & 0 & 0.5 \\ 0 & f & 0.5 \\ 0 & 0 & 1 \end{smallmatrix}\right]$, where the principal point is fixed at the normalized image center $(0.5, 0.5)$, and the focal length $f$ is randomly sampled from $(0.58, 0.88)$. For each pixel \( (u_i, v_i) \) in \( I_{\text{sin}} \), the corresponding 3D coordinates in the camera coordinate system are computed using \( K \) and the depth value \( D(u_i, v_i) \), thereby constructing the 3D point cloud \( P \):
\begin{equation}\label{equal2}
	P = \begin{bmatrix} X \\ Y \\ Z \end{bmatrix} = D(u_i, v_i) \cdot K^{-1} \begin{bmatrix} u_i \\ v_i \\ 1 \end{bmatrix}.
\end{equation}

Subsequently, we design a random camera extrinsic generator to enable switching to a novel view by sampling camera extrinsic parameters \([R\mid T]\). For the rotation component of extrinsic, the rotation angles are within $[-7.5^\circ, +7.5^\circ]$. For the translation component $t=[t_x,t_y,t_z]^\top$, the perturbation ranges are set to $[-0.3, +0.3]$ for $t_x$ and $t_y$, and $[-0.5, +0.5]$ for $t_z$. The data difficulty level can be modulated by adjusting the variation range of extrinsics. The point cloud \(P\) is transformed into the novel camera coordinate system via \(P' = R \cdot P + T\). Using differentiable rendering, \(P'\) is reprojected into the novel view to generate a preliminary rendered novel-view image \(I_{\text{novel}}\) and an occlusion mask \(M_{\text{occ}}\), while the Z-coordinates of \(P'\) directly constitute the corresponding depth map \(D'\) for \(I_{\text{novel}}\). To inpaint occluded regions in \(I_{\text{novel}}\) while preserving semantic continuity, we apply an image inpainting model \(M_{\text{inpaint}}\) for context-aware completion:
\begin{equation}\label{equal3}
	I_\text{inp} = M_\text{inpaint}(I_\text{novel}, M_\text{occ}).
\end{equation}

The view transformation module outputs a novel-view image \( I_{\text{inp}} \) with high 3D geometric fidelity, camera parameters \( \{K, [R \mid T]\} \) and depth maps \( \{D, D'\} \), which strictly follow 3D geometric consistency and circumvent the limitations of SfM. Using these outputs, we can establish the pixel-level GT correspondence between the synthetic image pair \( \{I_{\text{sin}}, I_{\text{inp}}\} \). In our current implementation, to maintain dense supervision, we uniformly treat all pixels (including inpainted ones) in the loss formulation. 
\subsection{Modality Transformation}
Another critical component in constructing a multi-modal image matching data\-base lies in leveraging a modality translation mechanism to transform the input single-view image into target modalities (\textit{e.g.}, infrared, depth, normal, and event), thereby simulating the modality differences inherent in real-world multi-modal image pairs, as shown in Fig.~\ref{fig1}. Specifically, for the original image \( I_{\text{sin}} \), multiple modality translation models are deployed in parallel to perform modality transformation, yielding a diverse set of pseudo-multi-modal images. In AnyMatch, we implement four representative target modalities: infrared, depth, normal, and event modalities.

For the \textbf{infrared} modality, we directly employ a pre-trained RGB-to-IR diffusion model. In the first stage, it collects a large-scale infrared image dataset for training. The text prompt for the diffusion model is uniformly set to \enquote{an infrared image}, and the Latent Diffusion Model (LDM) \cite{rombach2022high} is fine-tuned using Low-Rank Adaptation (LoRA) \cite{hu2022lora} to generate infrared images under text-conditioned guidance, thereby establishing a semantic association between the concept of \enquote{infrared} and its characteristic visual patterns. In the second stage, it collects a large-scale dataset of RGB-IR image pairs; during training, the RGB image serves as the conditional input to synthesize the corresponding infrared image, enabling the model to produce outputs that faithfully reflect the physical and perceptual characteristics of infrared imaging based on visible inputs.

For the \textbf{depth and normal} modalities, we directly employ pre-trained monocular relative depth estimation and normal estimation models, respectively.

For the \textbf{event} modality, we adopt a motion simulation method to synthesize event streams from static visible image pairs. Each pixel of the event camera independently monitors the logarithmic change in brightness. The logarithmic intensity is defined as $L = \log(I_{\text{sin}})$. When the change in brightness exceeds a temporal contrast threshold $C$, pixel $\mathbf{x}_k = (x_k, y_k)^\top$ triggers an event $e_k = (\mathbf{x}_k, t_k, p_k)$ at timestamp $t_k$. Therefore, the logarithmic change in brightness for each pixel is defined as $\Delta L(\mathbf{x}_k, t_k) = L(\mathbf{x}_k, t_k) - L(\mathbf{x}_k, t_k - \Delta t_k)$, where $\Delta L(\mathbf{x}_k, t_k) = p_k C$, with $C > 0$. $\Delta t_k$ is the time elapsed since the last triggered event for the same pixel, and $p_k = \pm 1$ represents the polarity of the brightness change (+1 for brightening, -1 for darkening). The threshold parameter $C \in [0.05, 0.5]$ and $p_k$ are set randomly to simulate different sensor characteristics. Furthermore, small random motions are applied to the input image frame $I_{sin}$ to simulate camera motion in real-world scenarios for calculating the event response.

The process of modality transformation can be simplified as:
\begin{equation}\label{equal4}
	I_\text{modal} = F(I_\text{sin}),
\end{equation}
where \( F \) denotes the modality transformation module, and \( I_{\text{modal}} \) represents the synthesized pseudo-target-modality image.
\subsection{Integration of View and Modality Transformation}
Since modality transformation preserves geometric structure without inducing distortions, we construct multi-modal and multi-view image pairs by combining the modality-transformed image $I_\text{modal}$ with the view-transformed image $I_\text{inp}$, while directly inheriting the GT supervision information generated during view transformation. Through the synergistic coupling of view and modality transformation, we ultimately synthesize the multi-modal and multi-view image matching dataset $T_\text{cross}$.
\subsection{Sample-level Geometric Consistency Verification}
During occlusion inpainting, image restoration may generate spurious content that violates the geometric and semantic consistency of the original scene—a phenomenon commonly termed \enquote{hallucination} \cite{liu2024survey} in generative vision literature. Although such content may appear locally coherent, it fundamentally deviates from the true scene structure and can adversely affect the learning of discriminative feature representations. To mitigate this issue, we introduce a geometric consistency verification mechanism for sample-level quality screening. As shown in Fig.~\ref{fig4}, for the single-modal image pair composed of the original image \( I_{\text{sin}} \) and the inpainted novel-view image \( I_{\text{inp}} \), we first employ a high-performance dense matching model to extract a dense correspondence set: 
\begin{figure*}[t]	
	\centering	
	\includegraphics[scale=0.124]{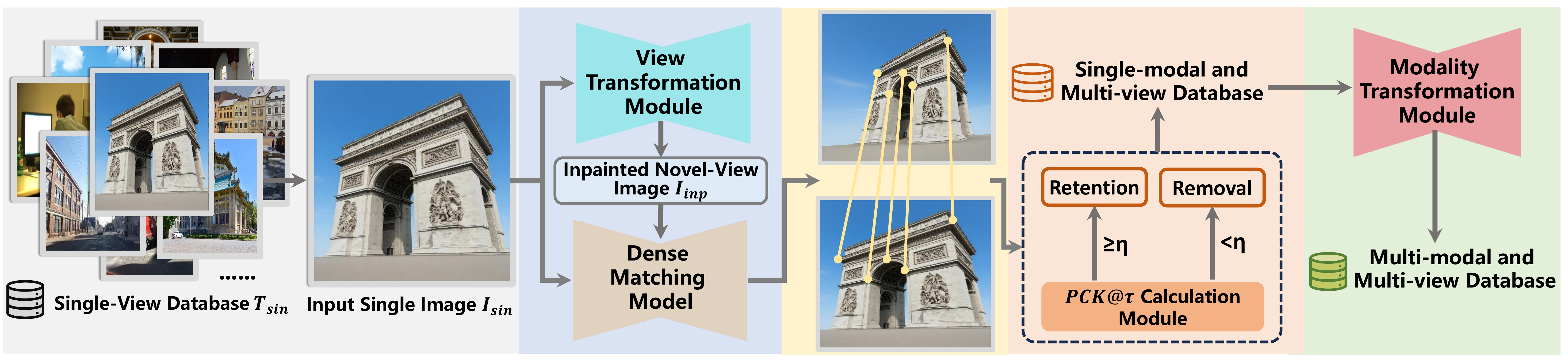}
	
	\caption{Overview of sample-level geometric consistency verification module. The image pair is retained if $\text{PCK}@\tau \geq \eta$ and removed if $\text{PCK}@\tau < \eta$.}\label{fig4}
	
\end{figure*}
\begin{equation}\label{equal5}
	\mathcal{M}_\text{init} = \{(x_\text{sin}^i, x_\text{inp}^i)\}_{i=1}^N,
\end{equation}
where $N$ denotes the total number of extracted correspondences. Subsequently, leveraging the known camera parameters and depth maps, the endpoint error (EPE) for each correspondence is computed as:
\begin{equation}\label{equal6}
	EPE(x_\text{sin}^i) = \left\| x_\text{inp}^i - \hat{x}_\text{inp}^i \right\|^2,
\end{equation}
where $x_{inp}^i$ is the predicted correspondence in $I_\text{inp}$ (from the matching model), and $\hat{x}_{inp}^i$ is the geometrically projected GT correspondence. The proportion $\text{PCK}@\tau$ of correct correspondences at threshold $\tau$ is defined as:
\begin{equation}\label{equal7}
	\text{PCK}@\tau = \frac{\left| \{(x_\text{sin}^i, x_\text{inp}^i)\}_{i=1}^N | EPE(x_\text{sin}^i) < \tau \right|}{N} \times 100\%.
\end{equation}

If $\text{PCK}@\tau$ of the synthesized image pair satisfies \( \text{PCK}@\tau \geq \eta \), the sample is retained as it demonstrates satisfactory geometric consistency and semantic coherence; otherwise, it is filtered out as an unreliable sample.

Through sample-level geometric consistency verification (SGCV), the quality of single-modal image pairs can be effectively optimized prior to modality transformation, thereby establishing a reliable foundation for generating trustworthy multi-modal data and ultimately enhancing the overall usability of the synthesized dataset. Based on the aforementioned pipeline, we construct a synthetic dataset named Any-syn, which comprises both training and testing subsets.

\section{Experiments}
\label{sec:Expe}
\subsection{Implementation Details}
\textbf{Training Strategy:} We directly adopt the official pre-trained models of EDM, LoFTR, and RoMa as initialization weights and perform fine-tuning on the Any-syn dataset. All models are trained on four NVIDIA RTX 4090 GPUs, with batch sizes of 4, 4, and 2 assigned to EDM, LoFTR, and RoMa, respectively. Tailored to the distinct characteristics of each baseline model, we employ different learning rates: for EDM, the original learning rate of $1 \times 10^{-4}$ is maintained throughout 10 epochs of fine-tuning; for LoFTR, a learning rate of $8 \times 10^{-4}$ is applied over 10 fine-tuning epochs; and for RoMa, the encoder and decoder learning rates are set to $6.25 \times 10^{-8}$ and $1.25 \times 10^{-6}$, respectively, achieving convergence within only 4 fine-tuning epochs.

\textbf{Model Details:} For AnyMatch, we employ Moge V1 \cite{wang2025moge} as the monocular relative depth estimation model, Moge V2 \cite{wang2026moge2} for surface normal map generation, and Stable Diffusion V2 \cite{rombach2022high} for image inpainting. The visible-to-infrared diffusion model is DiffV2IR \cite{ran2025diffv2ir}. The high-performance dense matching model used for geometric verification is RoMa \cite{edstedt2024roma}. For sample-level geometric consistency verification, the pixel threshold $\tau$ is set to $\tau= 5$ and the minimum correct match ratio $\eta$ is set to $\eta= 0.6$.
\begin{table*}[t]
	\scriptsize
	\renewcommand\arraystretch{0.8}  
	\centering
	\setlength{\belowcaptionskip}{3pt}
	\caption{Quantitative results (AUC of pose error) on the Any-syn dataset. The top-2 results of each category are highlighted as \colorbox{rankone}{first} and \colorbox{yellow!25}{second}.} 
	\setlength{\tabcolsep}{0.6mm}{
		\begin{tabular}{lcccccccccccc}			
			\hline
			\rule{0pt}{8pt}\multirow{2}{*}{\textbf{Method}} &\multicolumn{3}{c}{\textbf{RGB-IR}} &\multicolumn{3}{c}{\textbf{RGB-Depth}}&\multicolumn{3}{c}{\textbf{RGB-Normal}}&\multicolumn{3}{c}{\textbf{RGB-Event}} \\ 
			\cmidrule(r){2-4}\cmidrule(r){5-7}\cmidrule(r){8-10}\cmidrule(r){11-13}
			& @5$^\circ$ & @10$^\circ$ & @20$^\circ$ & @5$^\circ$ & @10$^\circ$ & @20$^\circ$ & @5$^\circ$ & @10$^\circ$ & @20$^\circ$ & @5$^\circ$ & @10$^\circ$ & @20$^\circ$\\	
			
			\hline	
			\rule{0pt}{8pt}\textbf{ReDFeat}
			&3.80& 10.64&23.19
			&0.20& 0.81&2.93
			&0.90&3.15&8.94
			& 0.19 &0.84&3.23 \\
			\textbf{XoFTR}
			&27.03&47.37&66.11
			&1.55& 4.05&9.29
			&10.33&21.62&35.49
			&0.89 & 2.34& 5.51
			\\
			\textbf{ELoFTR$_\text{PYM}$}
			&8.20 &18.51 &32.36
			&0.34 & 1.38&4.23
			&8.43&18.74&32.99
			& 1.15 &3.28&7.94
			\\
					
			\hline	
			\rule{0pt}{8pt}\textbf{EDM} &9.20&19.63&32.78&0.11&0.41&1.64&6.25&14.81&27.35&0.21&0.83&2.76   \\     
			\textbf{EDM$_\text{MINIMA}$} &\colorbox{ranktwo}{14.70}&\colorbox{ranktwo}{31.56}&\colorbox{ranktwo}{51.20}&\colorbox{ranktwo}{4.01}& \colorbox{ranktwo}{11.25}&\colorbox{ranktwo}{23.98}&\colorbox{ranktwo}{11.60}&\colorbox{ranktwo}{25.99}&\colorbox{ranktwo}{44.21}&\colorbox{ranktwo}{1.18}&\colorbox{ranktwo}{3.30}&\colorbox{ranktwo}{8.04}  \\  
			\rowcolor{gray!20} 
			\textbf{EDM$_\text{AnyMatch}$} &\colorbox{rankone}{28.80}&\colorbox{rankone}{50.00}&\colorbox{rankone}{68.53}&\colorbox{rankone}{8.32}&\colorbox{rankone}{20.36}&\colorbox{rankone}{38.09}&\colorbox{rankone}{18.23}&\colorbox{rankone}{36.65}&\colorbox{rankone}{56.60}&\colorbox{rankone}{3.31}&\colorbox{rankone}{9.32}&\colorbox{rankone}{19.86} \\ 
			\hline	
			\rule{0pt}{8pt}\textbf{LoFTR } &8.35&17.92&30.13&0.06&0.17&0.73&3.20&8.70&17.48&0.32&1.08&2.71 \\ 
			\textbf{LoFTR$_\text{MINIMA}$} &\colorbox{ranktwo}{13.09}&\colorbox{ranktwo}{28.75}&\colorbox{ranktwo}{48.20}&\colorbox{ranktwo}{2.29}&\colorbox{ranktwo}{9.27}&\colorbox{ranktwo}{19.98}&\colorbox{ranktwo}{9.89}&\colorbox{ranktwo}{22.59}&\colorbox{ranktwo}{39.96}&\colorbox{ranktwo}{1.06}&\colorbox{ranktwo}{3.43}&\colorbox{ranktwo}{8.67}\\
			\rowcolor{gray!20} 
			\textbf{LoFTR$_\text{AnyMatch}$} &\colorbox{rankone}{22.30}&\colorbox{rankone}{41.45}&\colorbox{rankone}{60.91}&\colorbox{rankone}{3.75}&\colorbox{rankone}{9.78}&\colorbox{rankone}{20.89}&\colorbox{rankone}{12.17}&\colorbox{rankone}{26.61}&\colorbox{rankone}{45.08}&\colorbox{rankone}{1.39}&\colorbox{rankone}{3.94}&\colorbox{rankone}{9.37}\\
			\hline	
			\rule{0pt}{8pt}\textbf{RoMa} &30.08&48.65&65.94&2.44&7.00&16.03&19.63&36.24&52.50&1.09&3.10&6.45\\
			\textbf{RoMa$_\text{MA}$} 
			&{34.45}&{55.68}&{72.82}
			&\colorbox{ranktwo}{10.22}&\colorbox{ranktwo}{22.76}&\colorbox{ranktwo}{39.21}
			&26.35&45.79&63.59
			&\colorbox{ranktwo}{3.42}&\colorbox{ranktwo}{9.16}&\colorbox{ranktwo}{18.31}
			\\
			\textbf{RoMa$_\text{MINIMA}$} &\colorbox{ranktwo}{40.00}&\colorbox{ranktwo}{60.62}&\colorbox{ranktwo}{76.27}&{9.76}&{22.16}&{38.59}&\colorbox{ranktwo}{29.97}&\colorbox{ranktwo}{49.33}&\colorbox{ranktwo}{65.81}&{2.52}&{7.29}&{14.84}\\
			\rowcolor{gray!20} 
			\textbf{RoMa$_\text{AnyMatch}$} &\colorbox{rankone}{50.91}&\colorbox{rankone}{69.88}&\colorbox{rankone}{82.75}&\colorbox{rankone}{14.05}&\colorbox{rankone}{29.39}&\colorbox{rankone}{47.39}&\colorbox{rankone}{36.28}&\colorbox{rankone}{56.50}&\colorbox{rankone}{72.23}&\colorbox{rankone}{4.81}&\colorbox{rankone}{11.87}&\colorbox{rankone}{22.13}\\
			
			\hline	
		\end{tabular}		
		\label{table1}	
		
	}	
\end{table*}
	
\textbf{Datasets:} We evaluate on five datasets, covering 17 cross-modal combinations. In the synthetic dataset Any-syn, the resolution of the images is $512\times 512$ pixels, covering five modal combinations of RGB-RGB/IR/Depth/Normal/Event. The synthetic training subset is constructed from the GLDv2 dataset \cite{weyand2020google}, comprising 500000 image pairs, while the synthetic test subset is derived from the SA-1B dataset \cite{kirillov2023segment}, containing 10000 image pairs. Following MINIMA, we train exclusively on three modality pairs: RGB-IR, RGB-Depth, and RGB-Normal. For real-world evaluation, METU-VisTIR \cite{tuzcuouglu2024xoftr} provides 2590 RGB-IR image pairs with pose annotations for pose estimation benchmarks. DIODE \cite{yang2024depth} contains 27858 pre-aligned RGB-Depth/Normal image pairs. DSEC \cite{wang2023visevent} supplies 100 rectified RGB-Event image pairs. To simulate geometric deformations while preserving evaluation fairness, synthetic homography transformations are applied to one image in each aligned pair prior to testing. Additionally, the MMIM dataset \cite{jiang2021review} encompasses 13 cross-modal combinations from remote sensing and medical imaging domains, with ground-truth correspondences manually annotated for homography transformation estimation tasks.

\textbf{Evaluation Protocol:} For pose estimation tasks, we report the Area Under the Curve (AUC) of the pose error at rotation error thresholds of {5$^\circ$, 10$^\circ$, 20$^\circ$}. For homography estimation tasks, we report the AUC of the maximum corner reprojection error across the four image corners at pixel thresholds of {3px, 5px, 10px}. To ensure fair comparison across methods, all robust geometric estimation procedures employ identical RANSAC \cite{fischler1981random} hyperparameters.
\subsection{Evaluation on Any-syn}
To validate the quality and learnability of Any-syn and the effectiveness of AnyMatch on synthetic data, we first conduct systematic pose estimation evaluations on the Any-syn test subset, comparing three model variants: the original pre-trained models, models fine-tuned by MINIMA, and models fine-tuned by AnyMatch. Furthermore, we also compare with ReDFeat, XoFTR, ELoFTR improved by PYM, and RoMa fine-tuned with MatchAnything (RoMa$_\text{MA}$). As shown in Tab.~\ref{table1}, models fine-tuned with AnyMatch consistently achieve superior performance across all multi-modal image pairs compared to baseline methods, MINIMA-fine-tuned counterparts, and other state-of-the-art multi-modal matchers, demonstrating that our framework effectively enhances cross-modal matching capabilities. Taking RoMa as an example, RoMa$_\text{AnyMatch}$ attains an AUC$@10^\circ$ of 69.88$\%$ on the RGB-IR task, representing an absolute improvement of 9.26$\%$ over RoMa$_\text{MINIMA}$ (60.62$\%$) and a substantial gain of 21.23$\%$ (43.6$\%$ relative improvement) over the original RoMa baseline (48.65$\%$). Owing to the fundamentally different sensing principle of event cameras—which record asynchronous per-pixel brightness changes as temporal event streams rather than full-frame intensities—the modality gap in RGB-Event pairs is significantly larger than in other modalities, resulting in generally lower performance across all methods. Nevertheless, even under this highly challenging setting, EDM$_\text{AnyMatch}$ achieves an AUC$@10^\circ$ of 9.32$\%$, substantially outperforming EDM$_\text{MINIMA}$ (3.30$\%$) by 6.02$\%$. These results collectively validate the efficacy of AnyMatch in generating geometrically consistent and semantically plausible synthetic data for robust cross-modal matching.
\begin{figure*}[t]	
	\centering	
	\includegraphics[scale=0.122]{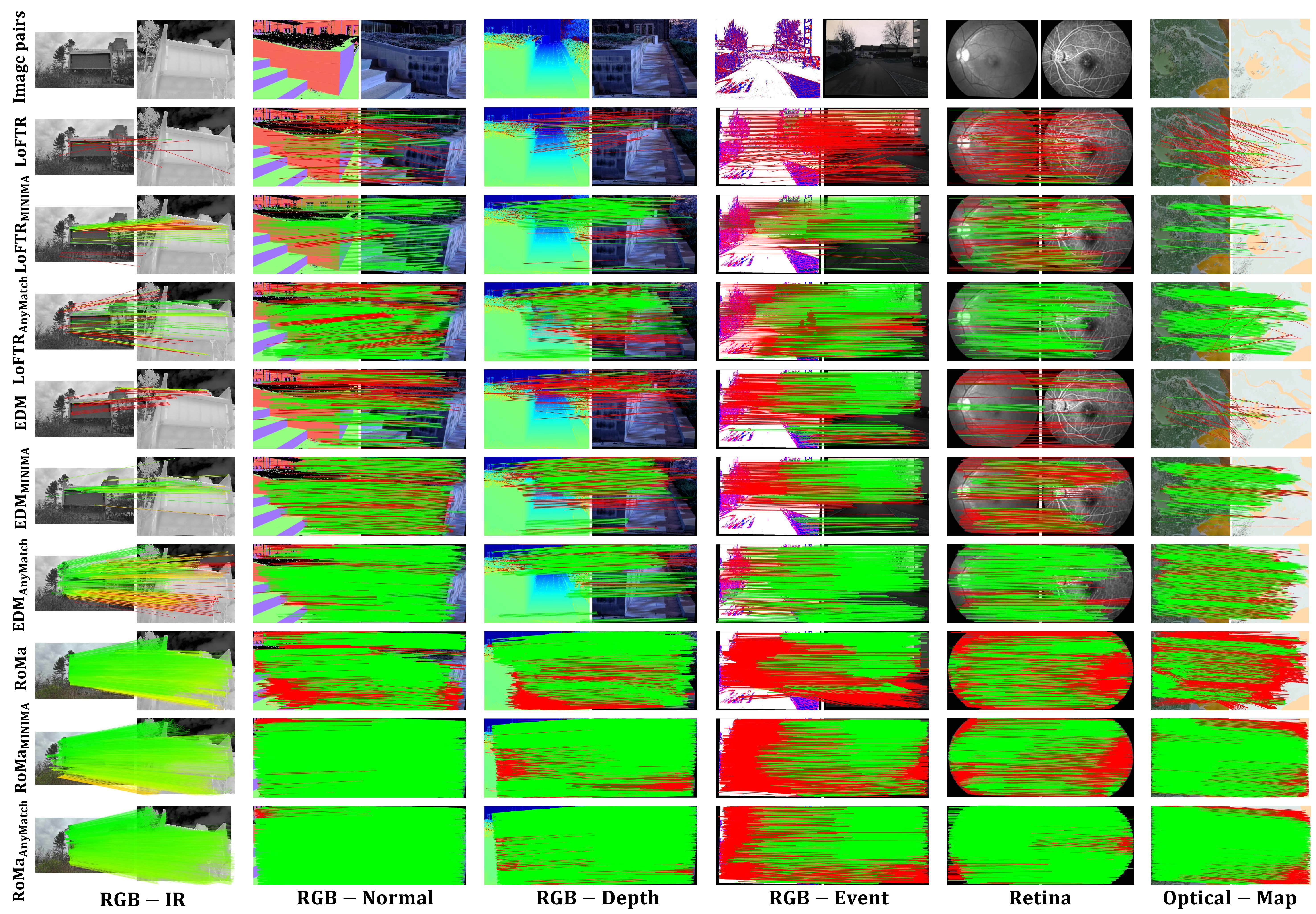}
	\caption{Qualitative results on real multi-modal images. Comparison of LoFTR, LoFTR$_\text{MINIMA}$, LoFTR$_\text{AnyMatch}$, EDM, EDM$_\text{MINIMA}$, EDM$_\text{AnyMatch}$, RoMa, RoMa$_\text{MINIMA}$, and RoMa$_\text{AnyMatch}$ in multiple image pairs, including RGB-IR, RGB-Normal, RGB-Depth, RGB-Event, Retina, and Optical-Map (from left to right). Matches generated by each method are drawn, where the red lines indicate epipolar error (pose) or projection error (homography) beyond $5 \times 10^{-4}$ or 3 pixels.}\label{fig3}
\end{figure*}
\subsection{Evaluation on Real Dataset}
\begin{table*}[t]
	\scriptsize
	\renewcommand\arraystretch{0.8}  
	\centering
	\setlength{\belowcaptionskip}{3pt}
	\caption{Quantitative results (AUC of pose error and reprojection error) in real RGB-IR/Depth/Normal/Event datasets. The top-2 results of each category are highlighted as \colorbox{rankone}{first} and \colorbox{ranktwo}{second}.}
	\setlength{\tabcolsep}{0.45mm}{
		\begin{tabular}{lcccccccccccc}			
			\hline
			\rule{0pt}{8pt}\multirow{2}{*}{\textbf{Method}} &\multicolumn{3}{c}{\textbf{RGB-IR}} &\multicolumn{3}{c}{\textbf{RGB-Depth}}&\multicolumn{3}{c}{\textbf{RGB-Normal}}&\multicolumn{3}{c}{\textbf{RGB-Event}} \\ 
			\cmidrule(r){2-4}\cmidrule(r){5-7}\cmidrule(r){8-10}\cmidrule(r){11-13}
			& @5$^\circ$ & @10$^\circ$ & @20$^\circ$ & @3px & @5px & @10px & @3px & @5px & @10px & @3px & @5px & @10px\\			
			\hline	
			\rule{0pt}{8pt}\textbf{ReDFeat}
			&1.71&4.57& 10.85
			&1.01& 4.58&16.30
			&0.19&1.59&8.71
			& 0.55& 0.97& 6.07\\
			\textbf{XoFTR}
			&18.47&34.64& 51.50
			&11.03& 27.24&51.60
			&19.79&40.52&64.47
			& 0.00& 1.37& 12.64
			\\
			\textbf{ELoFTR$_\text{PYM}$}
			&14.45 &29.73 & 46.55
			&25.70 & 42.90&62.45
			&1.75&7.02&21.37
			& {0.31}& 0.87& 10.64 
			\\
			\hline	
			\rule{0pt}{8pt}\textbf{EDM}
			&5.20&15.82&31.71
			&1.21&5.75&19.44
			&6.08&18.25&41.82
			&\colorbox{ranktwo}{0.60}&\colorbox{ranktwo}{1.60}&8.89   \\ 
			
			\textbf{EDM$_\text{MINIMA}$} 
			&\colorbox{ranktwo}{15.32}&\colorbox{ranktwo}{32.80}&\colorbox{ranktwo}{53.47}
			&\colorbox{ranktwo}{8.44}& \colorbox{ranktwo}{24.89}&\colorbox{ranktwo}{52.09}
			&\colorbox{ranktwo}{13.69}&\colorbox{ranktwo}{33.72}&\colorbox{ranktwo}{59.62}
			&{0.27}&{1.41}&\textbf{13.63}  \\  
			
			\rowcolor{gray!20} 
			\textbf{EDM$_\text{AnyMatch}$ }
			&\colorbox{rankone}{17.13}&\colorbox{rankone}{35.80}&\colorbox{rankone}{55.24}
			&\colorbox{rankone}{10.71}&\colorbox{rankone}{28.75}&\colorbox{rankone}{55.64}
			&\colorbox{rankone}{18.42}&\colorbox{rankone}{40.04}&\colorbox{rankone}{65.22}
			&\colorbox{rankone}{0.50}&\colorbox{rankone}{2.12}&\colorbox{ranktwo}{13.57} \\ 
			\hline	
			\rule{0pt}{8pt}\textbf{LoFTR } 
			&4.48&8.56&15.35
			&0.79&4.37&15.87
			&0.06&0.19&0.50
			&0.30&1.09&5.53 \\ 
			
			\textbf{LoFTR$_\text{MINIMA}$ }
			&\colorbox{rankone}{15.61}&\colorbox{rankone}{30.84}&\colorbox{rankone}{47.87}
			&\colorbox{ranktwo}{2.14}&\colorbox{ranktwo}{8.85}&\colorbox{ranktwo}{26.22}
			&\colorbox{rankone}{14.19}&\colorbox{ranktwo}{33.37}&\colorbox{ranktwo}{58.72}
			&\colorbox{rankone}{0.35}&\colorbox{ranktwo}{1.21}&\colorbox{ranktwo}{8.56}\\
			
			\rowcolor{gray!20} 
			\textbf{LoFTR$_\text{AnyMatch}$ }
			&\colorbox{ranktwo}{14.48}&\colorbox{ranktwo}{25.83}&\colorbox{ranktwo}{40.39}
			&\colorbox{rankone}{4.08}&\colorbox{rankone}{13.33}&\colorbox{rankone}{34.63}
			&\colorbox{ranktwo}{13.83}&\colorbox{rankone}{34.03}&\colorbox{rankone}{60.69}
			&\colorbox{ranktwo}{0.34}&\colorbox{rankone}{1.83}&\colorbox{rankone}{12.78}\\
			\hline	
			\rule{0pt}{8pt}\textbf{RoMa }
			&25.61&48.12&68.37
			&12.57&29.47&54.47
			&24.28&46.01&69.09
			&0.25&0.91&8.72\\
			
			\textbf{RoMa$_\text{MA}$ }
			&{30.42}&{52.77}&{71.60}
			&{26.33}&{48.80}&{71.07}
			&35.71&58.35&\colorbox{ranktwo}{78.37}
			&\colorbox{ranktwo}{0.26}&\colorbox{ranktwo}{1.51}&{7.66}
			\\
			
			\textbf{RoMa$_\text{MINIMA}$ }
			&\colorbox{ranktwo}{37.45}&\colorbox{ranktwo}{60.70}&\colorbox{rankone}{78.00}
			&\colorbox{rankone}{29.77}&\colorbox{rankone}{51.09}&\colorbox{rankone}{72.64}
			&\colorbox{ranktwo}{37.91}&\colorbox{ranktwo}{59.24}&{77.88}
			&{0.25}&{1.37}&\colorbox{ranktwo}{13.16}\\
			
			\rowcolor{gray!20} 
			\textbf{RoMa$_\text{AnyMatch}$} 
			&\colorbox{rankone}{40.49}&\colorbox{rankone}{62.61}&\colorbox{ranktwo}{77.76}
			&\colorbox{ranktwo}{29.12}&\colorbox{ranktwo}{49.83}&\colorbox{ranktwo}{71.50}
			&\colorbox{rankone}{39.26}&\colorbox{rankone}{60.02}&\colorbox{rankone}{78.35}
			&\colorbox{rankone}{0.32}&\colorbox{rankone}{1.64}&\colorbox{rankone}{14.25}\\
			\hline	
		\end{tabular}		
		\label{table2}	
	}	
\end{table*}	
\begin{table*}[t]
	\renewcommand
	\arraystretch{0.8}
	\caption{Quantitative results (AUC of reprojection error) in real medical and remote sensing dataset. The top-2 results of each category are highlighted as \colorbox{rankone}{first} and \colorbox{ranktwo}{second}.}
	\label{table3}
	\centering
	\setlength{\belowcaptionskip}{0.6pt}
	\scriptsize	
	\setlength{\tabcolsep}{3.5mm}{
		\begin{tabular}{lcccccc}		
			\hline
			\rule{0pt}{8pt}\multirow{2}{*}{\textbf{Method}} &\multicolumn{3}{c}{\textbf{Medical}} &\multicolumn{3}{c}{\textbf{Remote Sensing}
			} \\ 
			\cmidrule(r){2-4}\cmidrule(r){5-7}
			& @3px & @5px & @10px & @3px & @5px & @10px \\	
			
			\hline	
			\rule{0pt}{8pt}\textbf{ReDFeat}
			&31.44&39.16&45.62
			&1.92& 5.91&19.68\\
			\textbf{XoFTR}
			&38.97&45.23& 51.89
			&23.31&34.44&53.56
			\\
			\textbf{ELoFTR$_\text{PYM}$}
			&14.59 &21.90 & 32.15
			&0 &0 &10.12
			\\
			
			\hline	
			\rule{0pt}{8pt}\textbf{EDM }
			&38.79&44.10&50.61
			&16.95&26.50&44.12   \\ 
			
			\textbf{EDM$_\text{MINIMA}$ }
			&\colorbox{ranktwo}{39.57}&\colorbox{ranktwo}{45.42}&\colorbox{ranktwo}{52.09}
			&\colorbox{ranktwo}{18.29}& \colorbox{ranktwo}{30.52}&\colorbox{ranktwo}{52.81}  \\  
			
			\rowcolor{gray!20} 
			\textbf{EDM$_\text{AnyMatch}$} 
			&\colorbox{rankone}{39.76}&\colorbox{rankone}{45.79}&\colorbox{rankone}{55.18}
			&\colorbox{rankone}{25.17}&\colorbox{rankone}{38.69}&\colorbox{rankone}{57.90} \\ 
			\hline	
			\rule{0pt}{8pt}\textbf{LoFTR  }
			&37.89&42.73&47.36
			&5.94&9.34&12.57 \\ 
			
			\textbf{LoFTR$_\text{MINIMA}$ }
			&\colorbox{ranktwo}{38.86}&\colorbox{ranktwo}{44.79}&\colorbox{ranktwo}{52.35}
			&\colorbox{ranktwo}{20.93}&\colorbox{ranktwo}{33.96}&\colorbox{ranktwo}{54.30}\\
			
			\rowcolor{gray!20} 
			\textbf{LoFTR$_\text{AnyMatch}$} 
			&\colorbox{rankone}{38.97}&\colorbox{rankone}{45.17}&\colorbox{rankone}{52.93}
			&\colorbox{rankone}{24.36}&\colorbox{rankone}{39.76}&\colorbox{rankone}{58.77}\\
			\hline	
			\rule{0pt}{8pt}\textbf{RoMa }
			&\colorbox{rankone}{39.16}&\colorbox{ranktwo}{44.79}&53.89
			&26.29&37.55&55.45\\
			
			\textbf{RoMa$_\text{MA}$} 
			&{37.84}&{43.92}&{55.42}
			&{27.60}&{42.83}&{60.28}
			\\
			
			\textbf{RoMa$_\text{MINIMA}$} 
			&{37.92}&{44.67}&\colorbox{ranktwo}{57.07}
			&\colorbox{ranktwo}{29.27}&\colorbox{rankone}{44.19}&\colorbox{ranktwo}{63.08}\\
			
			\rowcolor{gray!20} 
			\textbf{RoMa$_\text{AnyMatch}$ }
			&\colorbox{ranktwo}{38.10}&\colorbox{rankone}{46.20}&\colorbox{rankone}{59.08}
			&\colorbox{rankone}{30.25}&\colorbox{ranktwo}{44.06}&\colorbox{rankone}{63.48}\\
			
			\hline	
	\end{tabular}}
\end{table*}

To validate the generalization capability of models trained by AnyMatch in real-world scenarios and provide more compelling evidence, we further conduct both in-domain and zero-shot evaluations across multiple real-world cross-modal datasets. Specifically, the RGB-IR dataset (METU-VisTIR) is employed for pose estimation benchmarks, while the remaining datasets (DIODE, DSEC, and MMIM) are utilized for homography estimation tasks.

\textbf{In-domain evaluation:} The results on real-world RGB-IR/RGB-Depth/ RGB-Normal datasets are presented in Tab.~\ref{table2} and Fig.~\ref{fig3}. For EDM, EDM$_\text{AnyMatch}$ consistently outperforms EDM$_\text{MINIMA}$ on nearly all modalities. Regarding RoMa and LoFTR, owing to the influence of inpainting hallucinations and systematic errors in depth estimation within the AnyMatch pipeline, MINIMA-fine-tuned models achieve marginally higher or comparable results on certain specific tasks (\textit{e.g.}, LoFTR on RGB-IR and RoMa on RGB-Depth). Nevertheless, AnyMatch maintains strong competitiveness against state-of-the-art multi-modal matchers.

\textbf{Zero-shot evaluation:} Zero-shot results on the unseen medical and remote sensing image datasets (MMIM) are reported in Tab.~\ref{table3} and Fig.~\ref{fig3}. In the results, AnyMatch demonstrates consistently superior generalization capability compared to MINIMA and other state-of-the-art matchers. 

AnyMatch not only delivers superior performance on known modalities but also establishes state-of-the-art results in zero-shot tasks involving unseen modalities. This demonstrates that models fine-tuned on AnyMatch learn modality-invariant feature representations and enhance 3D geometric awareness, achieving stronger generalization 
capability across diverse cross-modal matching scenarios.

\subsection{Ablation Studies}
To analyze the impact of various factors in AnyMatch on model performance, we adopt EDM as the baseline model and perform a series of ablation studies on the METU-VisTIR dataset.

\begin{wraptable}[10]{r}{0.547\textwidth} 
	\setlength{\intextsep}{0pt}   
	\setlength{\parskip}{0pt}
	\setlength{\abovecaptionskip}{-22pt}  
	\raggedleft
	\scriptsize
	\renewcommand\arraystretch{1}  
	\setlength{\tabcolsep}{1.3mm}
	\setlength{\belowcaptionskip}{8pt}
	\centering
	\caption{Ablation Studies. Test on real METU-VisTIR dataset using AUC of pose error.}
	\begin{tabular}{lccc}            
		\hline
		\rule{0pt}{8pt}{\textbf{Training Strategy}} &{\textbf{@5$^\circ$}} & {\textbf{@10$^\circ$}} & {\textbf{@20$^\circ$}} \\             
		\hline    
		\rule{0pt}{12pt}\textbf{1 modality w/o SGCV}&13.55&27.71&44.50  \\
		\textbf{3 modalities w/o SGCV}&15.63&31.24&48.29 \\
		\textbf{3 modalities w/ SGCV}&\colorbox{rankone}{17.13}&\colorbox{rankone}{35.80}&\colorbox{rankone}{55.24} \\
		\textbf{\makecell[l]{3 modalities w/ SGCV \\+ from scratch}} &5.69&14.16
		&27.45 \\  
		\textbf{\makecell[l]{3 modalities w/ SGCV \\+ homo}} &15.45&{30.78}&46.16 \\    
		\hline    
	\end{tabular}
	\label{table5} 
\end{wraptable}
(1) \textbf{Number of modalities.} As shown in Tab.~\ref{table5}, we fine-tune EDM using datasets containing either a single modality pair (RGB-IR) or three modality pairs (RGB-IR, RGB-Depth, RGB-Normal) to validate the impact of modality diversity on matching performance. The results demonstrate that joint training across multiple modalities consistently yields significant performance improvements compared to single-modality training, highlighting the benefit of multi-modal collaborative learning in enhancing cross-modal generalization.

(2) \textbf{Sample-level geometric consistency verification.} We apply the verification threshold $\eta$ with values of 0.7, 0.6, 0.5, 0.4, 0.3, and 0 (no filtering) to progressively filter the original synthetic dataset, and subsequently fine-tune EDM on each filtered subset. As reported in Tab.~\ref{table4}, the optimal performance is achieved when $\eta = 0.6$. This result demonstrates that selectively retaining samples with high geometric consistency effectively eliminates \enquote{hallucination} artifacts introduced during the generative process, thereby substantially enhancing model robustness and matching accuracy.

\begin{table}[t] 
	\raggedleft
	\scriptsize	
	\centering
	\renewcommand\arraystretch{1}  
	\setlength{\tabcolsep}{4mm}
	\setlength{\belowcaptionskip}{5pt}
	\caption{Results of sample-level geometric consistency verification using varying thresholds. Models fine-tuned on Any-syn subsets filtered by different verification thresholds ($\eta \in{\{0.7,0.6,0.5,0.4,0.3,0\}}$) are evaluated on the METU-VisTIR dataset.}
	\begin{tabular}{lcccccc}            
		\hline
		\rule{0pt}{8pt}\textbf{Threshold} & \textbf{0.7} & \textbf{0.6} & \textbf{0.5} & \textbf{0.4} & \textbf{0.3}&\textbf{0}\\             
		\hline    
		\rule{0pt}{10pt}\textbf{AUC@5$^\circ$}&15.50&\colorbox{rankone}{17.13}&15.39&15.36&15.73&15.63  \\
		\textbf{AUC@10$^\circ$}&32.52&\colorbox{rankone}{35.80}&31.47&32.04&31.32&31.24 \\
		\textbf{AUC@20$^\circ$}&51.67&\colorbox{rankone}{55.24}&48.97&50.08&48.55&48.29  \\    
		\hline    
	\end{tabular}
	\label{table4}    
\end{table}
(3) \textbf{Fine-tuning \textit{vs.} From Scratch.} Comparing the strategies of training from scratch and fine-tuning, results in Tab.~\ref{table5} reveal that models trained from scratch achieve substantially inferior performance compared to fine-tuned counterparts, with the AUC$@20^\circ$ metric reaching only approximately 50$\%$ of the fine-tuned model's score. This finding validates that leveraging the matching prior knowledge embedded in pre-trained models is crucial for effective convergence and robust adaptation in cross-modal matching tasks.

(4) \textbf{2D Homography-based Synthesis \textit{vs.} 3D View Transformation-based Synthesis.} We construct a homography-synthetic dataset (homo-syn) by applying synthetic homography transformations to original RGB images followed by the same modality transformation pipeline in AnyMatch. In Tab.~\ref{table5}, models fine-tuned on homo-syn exhibit substantially degraded performance compared to those fine-tuned on Any-syn. This result validates that disparity and occlusion relationships generated via 3D reprojection better conform to real-world physical constraints and geometric priors, thereby yielding more realistic and geometrically consistent training data than simplistic planar transformations.

\section{Conclusion}
In conclusion, to address the scarcity of large-scale training data with precise geometric annotations for multi-modal image matching, we propose AnyMatch—a framework that transforms abundant, easily accessible single-view images at minimal cost into multi-modal training data with high 3D geometric fidelity. By decoupling view and modality transformation, AnyMatch integrates monocular depth estimation, 3D reprojection, diffusion-based inpainting, and cross-modal translation to synthesize multi-view, multi-modal pairs and annotations that adhere to 3D geometric consistency, while employing Sample-level Geometric Consistency Verification (SGCV) to filter generative artifacts. Moreover, AnyMatch offers strong scalability, enabling controllable scene diversity and annotation difficulty levels via adjustable input and camera parameters. Experiments on the Any-syn dataset demonstrate that matching networks (LoFTR, EDM, RoMa) fine-tuned on Any-syn achieve substantial gains on cross-modal benchmarks, exhibiting superior generalization and robustness over baselines and MINIMA-fine-tuned models, including in zero-shot evaluations on unseen modalities.
\section*{Acknowledgements}
This work was supported by the National Natural Science Foundation of China (Grant Nos. 62276192 and 62475199) and National Key R\&D Program of China (Grant No. 2024YFE0202500).

\bibliographystyle{splncs04}
\bibliography{main}
\clearpage
\appendix
\setcounter{section}{0}
\setcounter{figure}{0}
\setcounter{table}{0}
\setcounter{equation}{0}
\renewcommand{\thefigure}{S\arabic{figure}}
\renewcommand{\thetable}{S\arabic{table}}
\renewcommand{\theequation}{S\arabic{equation}}
\makeatletter
\@ifpackageloaded{hyperref}{%
	\renewcommand{\theHsection}{supp.\arabic{section}}
	\renewcommand{\theHfigure}{supp.\arabic{figure}}
	\renewcommand{\theHtable}{supp.\arabic{table}}
	\renewcommand{\theHequation}{supp.\arabic{equation}}
}{}
\makeatother
\section*{Supplementary Material}
\addcontentsline{toc}{section}{Supplementary Material}
\section{Details of View Transformation}
\label{supp:sec:intro}
The view transformation module in AnyMatch comprises two key components: 3D point cloud reconstruction and differentiable mesh rendering. This method aims to recover dense 3D geometric structure from a single RGB image, construct a triangular mesh model, and perform re-rendering under randomly sampled camera extrinsics, thereby generating novel-view images with 3D geometric consistency.

\subsection{3D Point Cloud Construction}
As detailed in the main paper, given an input image $I_{\text{sin}}$, we first employ a pre-trained monocular depth estimation model $M_{\text{rd}}$ to predict an initial depth map $D_r$, which undergoes random affine perturbations to yield $D$. Subsequently, the pixel coordinates of $I_{\text{sin}}$ are projected into the camera coordinate system to construct a 3D point cloud $P$, utilizing randomly generated normalized camera intrinsics $K$.

\subsection{Differentiable Mesh Rendering}
To support differentiable rendering, the dense point cloud is converted into a triangular mesh $\mathcal{M}=(\mathcal{V}, \mathcal{F})$. The vertex set $\mathcal{V}$ comprises the flattened point cloud $P$. The face set $\mathcal{F}$ is generated based on the image pixel grid, where each pixel cell $(i, j)$ is subdivided into two triangles, with vertex indices following a regular grid topology. Each vertex is associated with an attribute vector $\mathbf{A}=[{c}, m, o]^\top$, containing color ${c}$, a visibility mask $m$, and an object mask $o$. The visibility mask is calculated based on the gradient magnitude of the disparity map, utilized to cull unreliable vertices in regions of depth discontinuity.

To synthesize novel view, the vertices are transformed into the target camera coordinate system. Given randomly sampled camera extrinsics ${T}_{\text{ext}}$, the vertex set $\mathcal{V}$ is transformed into the target camera space to yield ${\mathcal{V}}_{\text{world}}$:
\begin{equation}\label{supp:equal1}
{\mathcal{V}}_\text{world} = {T}_\text{ext} \cdot {\mathcal{V}}^{\top}.
\end{equation}

This step establishes the mathematical relation between the 3D geometric structure and the target view pose.

To project the 3D scene into the 2D Normalized Device Coordinates (NDC) space, a perspective projection matrix $\mathbf{\Pi} \in \mathbb{R}^{4 \times 4}$ is constructed. $\mathbf{\Pi} $ is determined by the camera intrinsic ${K}$ and the clipping plane parameters. Specifically, let the focal lengths be $f_x, f_y$, the principal point coordinates be $c_x, c_y$, and the near and far clipping planes be $z_{\text{near}}$ and $z_{\text{far}}$, respectively. The projection matrix is formulated as follows:
\begin{equation}\label{supp:equal2}
	\mathbf{\Pi} = \begin{bmatrix}
		2f_x & 0 & 2c_x - 1 & 0 \\
		0 & 2f_y & 2c_y - 1 & 0 \\
		0 & 0 & \frac{z_{near} + z_{far}}{z_{far} - z_{near}} & \frac{-2z_{near}z_{far}}{z_{far} - z_{near}} \\
		0 & 0 & 1 & 0
	\end{bmatrix}.
\end{equation}

$\mathbf{\Pi} $ projects the coordinates ${\mathcal{V}}_{\text{world}}$ from the target camera space into the clip space ${\mathcal{V}}_{\text{clip}}$:
\begin{equation}\label{supp:equal3}
	{\mathcal{V}}_\text{clip} = \mathbf{\Pi} \cdot {\mathcal{V}}_\text{world}^{\top}.
\end{equation}

Subsequently, the clip coordinates are transformed into NDC coordinates $\mathcal{V}_{\text{ndc}} \in \mathbb{R}^{B \times N \times 3}$ via perspective division:
\begin{equation}\label{supp:equal4}
	\mathcal{V}_\text{ndc}^{xyz} = \frac{{\mathcal{V}}_\text{clip}^{xyz}}{{\mathcal{V}}_\text{clip}^{w}}.
\end{equation}

Following perspective projection and coordinate transformation, the triangular mesh faces are mapped into the pixel coordinate system. For each pixel $(u, v)$, the renderer determines whether it lies within a projected triangle. If a pixel is covered by a face, the associated face index and the pixel's relative position within the triangle (barycentric coordinates $\boldsymbol{\lambda}=[\lambda_1,\lambda_2,\lambda_3]^\top$) are recorded. This process establishes a sparse correspondence between 2D pixels and 3D geometric structures, while generating a visibility mask to identify valid rendering regions. Additionally, the transformed target depth is concatenated as a component of the vertex attribute vector.

Once the correspondences between pixels and triangular faces are established, pixel attributes (\textit{e.g.}, color and disparity) are computed via weighted interpolation of the attributes from the three vertices of the corresponding triangular face. Denoting the attribute vectors of the three vertices as $\mathbf{A}_1, \mathbf{A}_2, \mathbf{A}_3$, the pixel attribute $\mathbf{A}_{\text{pixel}}$ is formulated as:
\begin{equation}\label{supp:equal5}
	\mathbf{A}_{pixel} = \sum_{i=1}^{3} \lambda_i \mathbf{A}_i, \text{ s.t. } \sum_{i=1}^{3} \lambda_i = 1,
\end{equation}
where $\boldsymbol{\lambda}$ represents the barycentric coordinates of the vertices. As $\boldsymbol{\lambda}$ depends on the projected positions of the vertices in screen space, and these projections are continuous functions of the 3D vertex coordinates, the interpolation process is inherently differentiable. Consequently, the renderer produces interpolated color and depth maps, which are combined with a visibility mask to exclude background regions, yielding geometrically consistent novel-view images.
\section{Data Synthesis Results of AnyMatch}
We execute the AnyMatch pipeline on the GLDv2 dataset to obtain the training subset of the Any-syn dataset. In the synthetic dataset Any-syn, the resolution of the images is $512\times 512$ pixels, covering five modal combinations of RGB-RGB/IR/Depth/Normal/Event. In Fig.~\ref{supp:fig1} and Fig.~\ref{supp:fig2}, we present some data synthesis results from the Any-syn dataset.

\section{Additional Qualitative Results}
We further show additional qualitative results selected from real-world datasets, including RGB-IR, RGB-Depth, RGB-Normal, RGB-Event, medical (retina), and remote sensing (including day-night, optical-infrared, and optical-map). Red lines denote false matches whose epipolar errors (for pose estimation) or projection errors (for homography estimation) exceed $5 \times 10^{-4}$ and 3 pixels, respectively. Fig.~\ref{supp:fig3}, Fig.~\ref{supp:fig4}, Fig.~\ref{supp:fig5}, Fig.~\ref{supp:fig6} and Fig.~\ref{supp:fig7} show the qualitative results of LoFTR, LoFTR$_{\text{MINIMA}}$, LoFTR$_{\text{AnyMatch}}$, EDM, EDM$_{\text{MINIMA}}$, EDM$_{\text{AnyMatch}}$, RoMa, RoMa$_{\text{MINIMA}}$, and RoMa$_{\text{AnyMatch}}$. The results demonstrate that our AnyMatch can yield a high quantity and proportion of correct matches (indicated by green lines).
\begin{table*}[t]
	\scriptsize
	\renewcommand\arraystretch{1}  
	\centering
	\setlength{\belowcaptionskip}{3pt}
	\caption{Quantitative results in real RGB-IR/Depth/Normal/Event datasets. The top-2 results of each category are highlighted as \colorbox{rankone}{first} and \colorbox{yellow!25}{second}.}
	\setlength{\tabcolsep}{0.4mm}{
		\begin{tabular}{lcccccccccccc}			
			\hline
			\rule{0pt}{8pt}\multirow{2}{*}{\textbf{Method}} &\multicolumn{3}{c}{\textbf{RGB-IR}} &\multicolumn{3}{c}{\textbf{RGB-Depth}}&\multicolumn{3}{c}{\textbf{RGB-Normal}}&\multicolumn{3}{c}{\textbf{RGB-Event}} \\ 
			\cmidrule(r){2-4}\cmidrule(r){5-7}\cmidrule(r){8-10}\cmidrule(r){11-13}
			& @5$^\circ$ & @10$^\circ$ & @20$^\circ$ & @3px & @5px & @10px & @3px & @5px & @10px & @3px & @5px & @10px\\			
			\hline	
			\rule{0pt}{8pt}\textbf{EDM }
			&5.20&15.82&31.71
			&1.21&5.75&19.44
			&6.08&18.25&41.82
			&\colorbox{ranktwo}{0.60}&\colorbox{ranktwo}{1.60}&8.87   \\ 
			
			\textbf{EDM$_\text{MINIMA}$ }
			&{15.32}&\colorbox{ranktwo}{32.80}&\colorbox{ranktwo}{53.47}
			&\colorbox{ranktwo}{8.44}& \colorbox{ranktwo}{24.89}&\colorbox{ranktwo}{52.09}
			&\colorbox{ranktwo}{13.69}&\colorbox{ranktwo}{33.72}&\colorbox{ranktwo}{59.62}
			&{0.27}&{1.41}&\textbf{13.63}  \\ 
			
			\textbf{EDM$_\text{homo}$ }
			&\colorbox{ranktwo}{15.45}&{30.78}&{46.16}
			&{3.47}&{11.74}&{31.51}
			&{9.90}&{26.99}&{52.84}
			&{0.43}&{1.33}&{7.26}  \\  
			
			\rowcolor{gray!20} 
			\textbf{EDM$_\text{AnyMatch}$ }
			&\colorbox{rankone}{17.13}&\colorbox{rankone}{35.80}&\colorbox{rankone}{55.24}
			&\colorbox{rankone}{10.71}&\colorbox{rankone}{28.75}&\colorbox{rankone}{55.64}
			&\colorbox{rankone}{18.42}&\colorbox{rankone}{40.04}&\colorbox{rankone}{65.22}
			&\colorbox{rankone}{0.50}&\colorbox{rankone}{2.12}&\colorbox{ranktwo}{13.57} \\ 
			\hline	
		\end{tabular}		
		\label{supp:table1}	
	}	
\end{table*}	
\begin{table}[t]
	\renewcommand
	\arraystretch{0.9}
	\caption{Quantitative results in real medical and remote sensing dataset. The top-2 results of each category are highlighted as \colorbox{rankone}{first} and \colorbox{yellow!25}{second}.}
	\label{supp:table2}
	\centering
	\setlength{\belowcaptionskip}{1pt}
	\small
	\scriptsize	
	\setlength{\tabcolsep}{2.8mm}{
		\begin{tabular}{lcccccc}		
			\hline
			\rule{0pt}{8pt}\multirow{2}{*}{\textbf{Method}} &\multicolumn{3}{c}{\textbf{Medical}} &\multicolumn{3}{c}{\textbf{Remote Sensing}
			} \\ 
			\cmidrule(r){2-4}\cmidrule(r){5-7}
			& @3px & @5px & @10px & @3px & @5px & @10px \\			
			\hline	
			\rule{0pt}{8pt}\textbf{EDM} 
			&38.79&44.10&50.61
			&16.95&26.50&44.12   \\ 
			
			\textbf{EDM$_\text{MINIMA}$} 
			&{39.57}&\colorbox{ranktwo}{45.42}&{52.09}
			&\colorbox{ranktwo}{18.29}& {30.52}&\colorbox{ranktwo}{52.81}  \\  
			
			\textbf{EDM$_\text{homo}$} 
			&\colorbox{ranktwo}{39.63}&{45.40}&\colorbox{ranktwo}{52.68}
			&{18.20}& \colorbox{ranktwo}{30.99}&{48.72}  \\
			
			\rowcolor{gray!20} 
			\textbf{EDM$_\text{AnyMatch}$ }
			&\colorbox{rankone}{39.76}&\colorbox{rankone}{45.79}&\colorbox{rankone}{55.18}
			&\colorbox{rankone}{25.17}&\colorbox{rankone}{38.69}&\colorbox{rankone}{57.90} \\ 
			
			\hline	
	\end{tabular}}
	
\end{table}
\section{Quantitative Results of 2D Homography-based Synthesis Method}
We show the quantitative results of the 2D Homography-based Synthesis method on real-world RGB-IR, RGB-Depth, RGB-Event, medical, and remote sensing datasets in Tab.~\ref{supp:table1} and Tab.~\ref{supp:table2}. Since this method employs planar homography transformations to simulate geometric deformations while neglecting 3D geometric consistency across views (\textit{e.g.}, the physical relationship between disparity and depth), its overall matching performance is inferior to that of both MINIMA and AnyMatch.
\begin{table}[t]
	\scriptsize
	\renewcommand\arraystretch{1.2}  
	\centering
	\setlength{\belowcaptionskip}{3pt}
	\caption{Quantitative results of LoFTR$_\text{AnyMatch}$ (black) and EDM$_\text{AnyMatch}$ (\textcolor{cyan}{cyan}) in three groups of RGB-IR. The values in parentheses are the percentage decrease relative to the third row.}
	\setlength{\tabcolsep}{4mm}{
		\begin{tabular}{lcccccc}			
			\hline
			\rule{0pt}{8pt}\textbf{\text{PCK}@$\tau$}& \textbf{AUC@5$^\circ$} & \textbf{AUC@10$^\circ$} &\textbf{AUC@20$^\circ$}   \\			
			\hline	
			\rule{0pt}{12pt}\textbf{[0.2,0.4]} 
			&\makecell{17.73($\downarrow$37.75\%)\\\textcolor{cyan}{28.87($\downarrow$15.44\%)} }
			&\makecell{32.99($\downarrow$30.94\%)\\ \textcolor{cyan}{48.61($\downarrow$10.84\%)}}
			&\makecell{48.76($\downarrow$24.96\%)\\ \textcolor{cyan}{66.39($\downarrow$6.31\%)}}
			
			\\
			
			\textbf{[0.4,0.6]}
			&\makecell{{23.22}($\downarrow$18.47\%)\\\textcolor{cyan}{30.83($\downarrow$9.700\%)}}
			&\makecell{{42.04}($\downarrow$12.00\%)\\\textcolor{cyan}{51.85($\downarrow$4.900\%)}}
			&\makecell{{59.49}($\downarrow$8.45\%)\\\textcolor{cyan}{69.45($\downarrow$1.99\%)}}
			
			\\
			
			\textbf{[0.6,0.8]} 
			&\makecell{{28.48}\\\textcolor{cyan}{34.14}}
			&\makecell{{47.77}\\\textcolor{cyan}{54.52}}
			&\makecell{{64.98}\\\textcolor{cyan}{70.86}}
			
			\\
			
			\hline	
		\end{tabular}		
		\label{supp:table3}	
	}	
\end{table}	
\section{Analysis of LoFTR's Performance Degradation}
To investigate why LoFTR$_\text{AnyMatch}$ underperforms LoFTR$_\text{MINIMA}$ on real RGB‑IR data despite our geometrically superior annotations, we conducted experiments for analysis.
As shown in Tab.~\ref{supp:table3}, we use the \text{PCK}@$\tau$ to partition the RGB-IR image pairs of Any-syn into three different quality groups (low [0.2,0.4], medium [0.4,0.6], and high [0.6,0.8]), and evaluate LoFTR$_\text{AnyMatch}$ and EDM$_\text{AnyMatch}$ on them. As the image quality becomes worse, the performance of matchers also degrades, and LoFTR degrades more strongly than EDM.
This suggests that LoFTR is more sensitive to hallucination/depth errors in lower-quality synthetic pairs. Thus the issue is a combination of imperfect synthetic data and matcher sensitivity.
\begin{table}[t]
	\raggedleft
	\scriptsize
	\renewcommand\arraystretch{1}  
	\setlength{\tabcolsep}{1mm}
	\setlength{\belowcaptionskip}{10pt}
	\centering
	\caption{Ablation Studies of RoMa on the METU-VisTIR dataset.}
	\setlength{\tabcolsep}{4mm}{
		\begin{tabular}{lccc}            
			\hline
			\rule{0pt}{8pt}\textbf{Training Strategy} &\textbf{@5$^\circ$} & \textbf{@10$^\circ$} & \textbf{@20$^\circ$} \\             
			\hline    
			\rule{0pt}{7pt}\textbf{3 modalities w/o SGCV}&33.07&54.56&71.75 \\
			\textbf{3 modalities w/ SGCV}&\colorbox{rankone}{40.49}&\colorbox{rankone}{62.61}&\colorbox{rankone}{77.76} \\   
			\textbf{3 modalities w/ SGCV + homo} &28.44&{51.10}&70.18 \\    
			\hline    
	\end{tabular}}
	\label{supp:table4} 
\end{table}
\section{Ablation Study on RoMa}
To verify that our key conclusions from the EDM‑based ablations (Tables 4 in the main paper) are not architecture‑specific, we repeat the two most critical experiments, namely sample‑level geometric consistency verification (SGCV) and 2D homography-based synthesis \textit{vs.} 3D view transformation-based synthesis, using RoMa as the backbone matcher. Tab.~\ref{supp:table4} shows the ablation study of RoMa on the METU-VisTIR dataset, which has the same trend as EDM, demonstrating that our conclusions, especially the effectiveness of SGCV and the superiority of 3D view synthesis over homography, are robust and generalise to other state‑of‑the‑art matchers.
\begin{table}[htbp] 
	\centering 
	\scriptsize
	\caption{Quantitative results of EDM training under varying focal lengths and relative camera transformations in RGB-IR dataset.}
	\begin{minipage}{0.4\textwidth}
		\setlength{\tabcolsep}{2.3mm}{
			\begin{tabular}{lccc}			
				\hline
				\rule{0pt}{8pt}\textbf{Focal Length}& \textbf{@5$^\circ$} & \textbf{@10$^\circ$} & \textbf{@20$^\circ$}  \\			
				\hline	
				\rule{0pt}{8pt}\textbf{[0.58,0.68]} 
				&29.86&51.54&70.41
				\\
				
				\textbf{[0.68,0.78]}
				&{29.72}&{50.05}&{68.39}
				\\
				
				\textbf{[0.78,0.88]} 
				&{27.37}&{48.15}&{67.66}
				\\
				
				\hline	
		\end{tabular}		}
	\end{minipage}
	\hspace{1.cm} 
	\begin{minipage}{0.48\textwidth}
		\centering
		\setlength{\tabcolsep}{2.3mm}{
			\begin{tabular}{lccc}			
				\hline
				\rule{0pt}{8pt}\textbf{Overlap Ratio}& \textbf{@5$^\circ$} & \textbf{@10$^\circ$} & \textbf{@20$^\circ$} \\			
				\hline	
				\rule{0pt}{8pt}\textbf{[0.0,0.4]} 
				&26.24&47.19&66.70
				\\
				
				\textbf{[0.4,0.7]}
				&{30.61}&{52.87}&{71.80}
				\\
				
				\textbf{[0.7,1.0]} 
				&{50.27}&{68.73}&{79.40}
				\\
				
				\hline	
			\end{tabular}		
		}	
	\end{minipage}
	\label{supp:table5}	
\end{table}
\section{Sensitivity to Camera Parameter}
To validate the rationality of our randomly sampled camera intrinsic and extrinsic ranges and to assess their impact on downstream matching performance, we conduct a sensitivity analysis using EDM as the baseline matcher. Specifically, we keep all other parameters fixed and partition the focal length into three intervals: [0.58,0.68], [0.68,0.78], [0.78,0.88]. The Any‑syn training subset is then divided into three groups according to the focal length, and EDM is trained separately on each group. Evaluation is performed on the RGB‑IR image pairs of Any‑syn test subset. As shown in the left of Tab.~\ref{supp:table5}, the model achieves the best performance with the intermediate focal range [0.58,0.68], whereas the performance drops moderately when shifting to higher focal ranges. This indicates that our chosen focal range is reasonable and that the model remains robust within this interval.

Similarly, we examine the influence of relative camera transformations by grouping Any‑syn training subset according to the overlap ratio between the original and novel views. The results in the right of Tab.~\ref{supp:table5} reveal a clear monotonic trend: performance improves steadily as overlap increases, with the high‑overlap group ([0.7,1.0]) achieving the best results, while the low‑overlap group ([0.0,0.4]) lags substantially behind. This confirms that the synthesized data exhibit physically plausible geometric relationships, and the model's performance correlates reasonably with the degree of viewpoint variation. Overall, these analyses demonstrate that our parameter choices are both practical and effective, and they provide useful guidance for selecting appropriate data difficulty levels in future applications.

\section{Results of Sample-level Geometric Consistency Verification}
\begin{figure*}[h]	
	\centering	
	\includegraphics[scale=0.35]{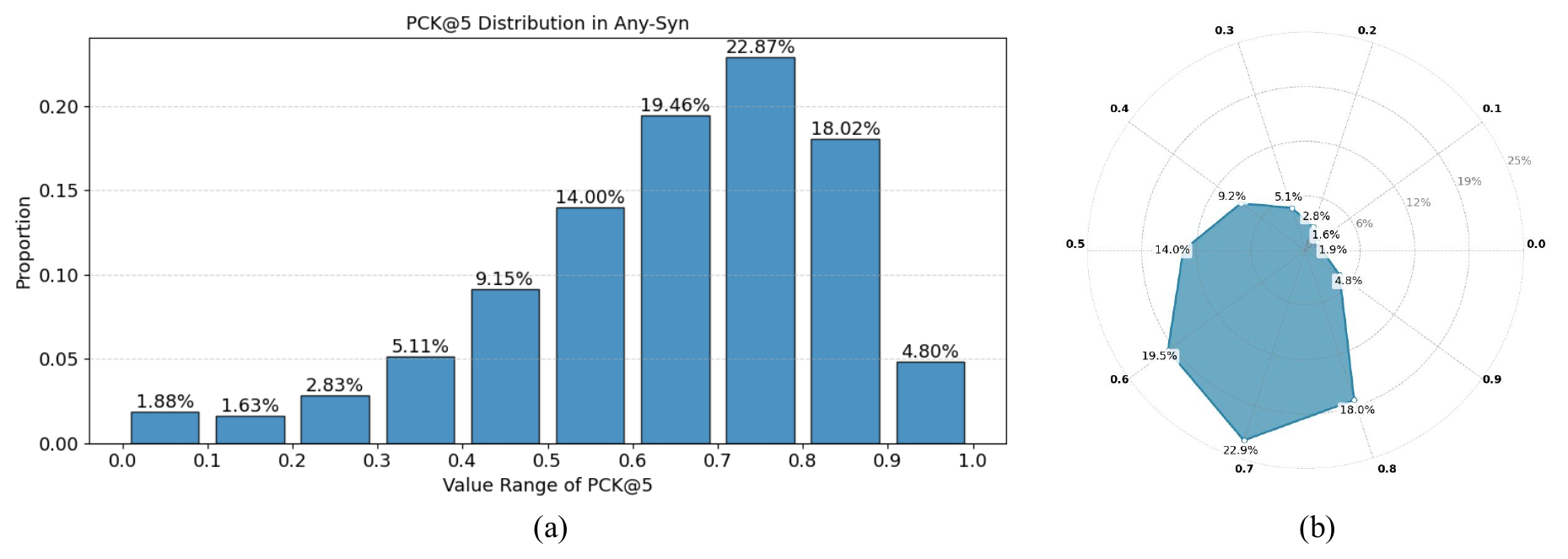}
	\caption{Distribution of PCK@5 in the Any-Syn dataset. (a) Histogram distribution; (b) Radar chart distribution.}\label{supp:fig12}
\end{figure*}

\begin{figure*}[h]	
	\centering	
	\includegraphics[scale=0.35]{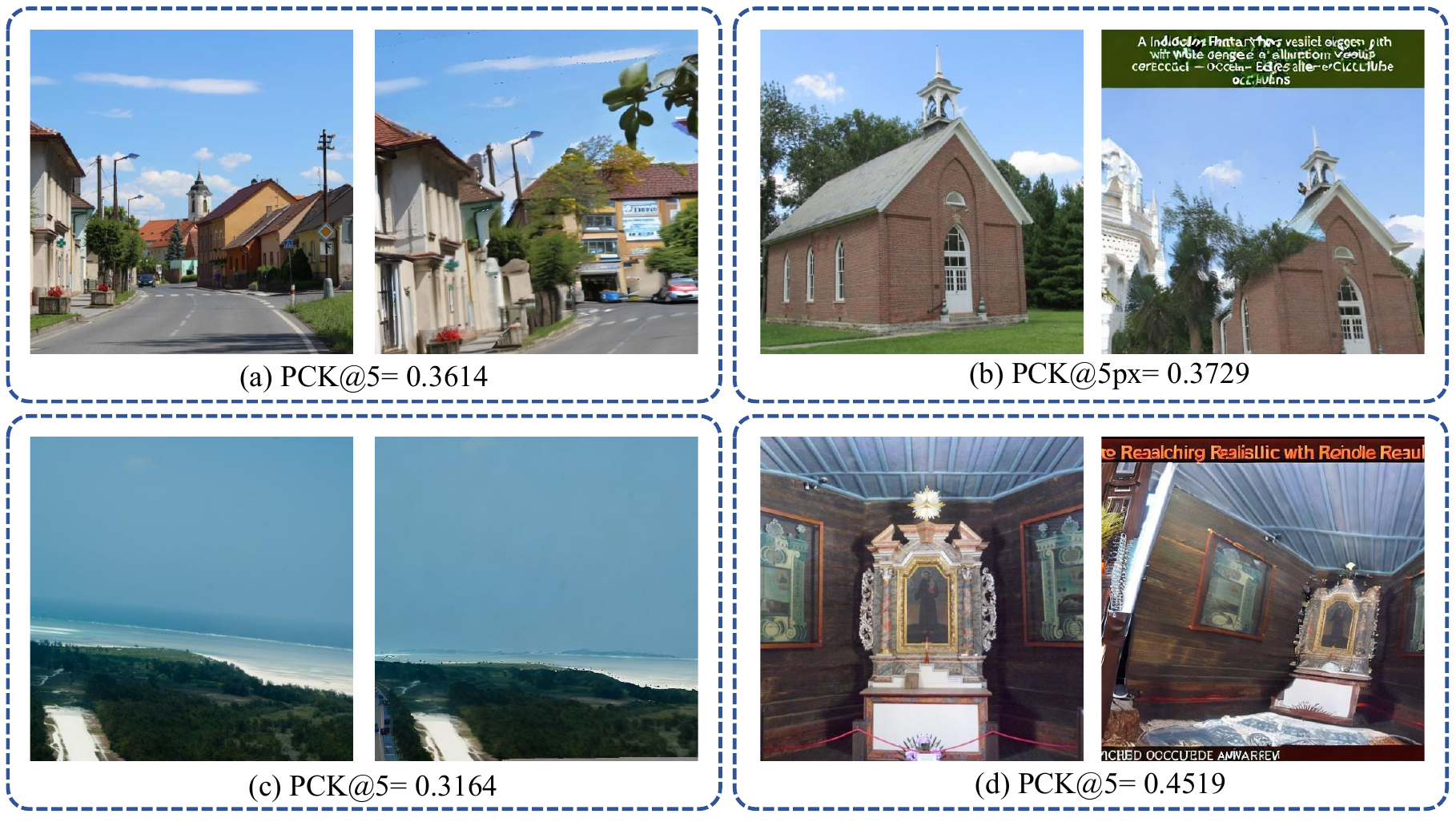}
	\caption{Image pairs with relatively high PCK@5.}\label{supp:fig13}
\end{figure*}

\begin{figure*}[h]	
	\centering	
	\includegraphics[scale=0.35]{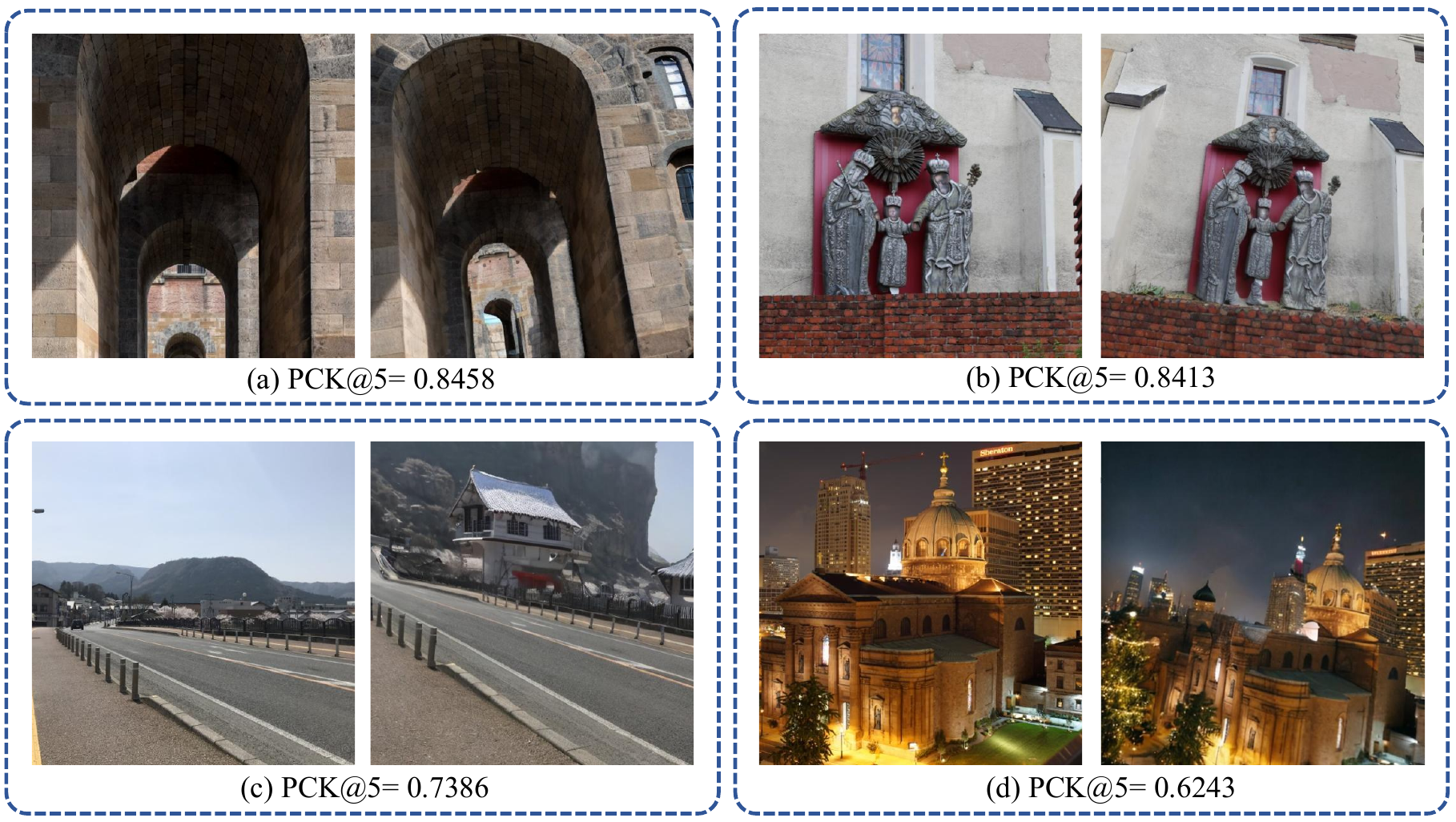}
	\caption{Image pairs with relatively low PCK@5.}\label{supp:fig14}
\end{figure*}

\begin{figure*}[t]	
	\centering	
	\includegraphics[scale=0.144]{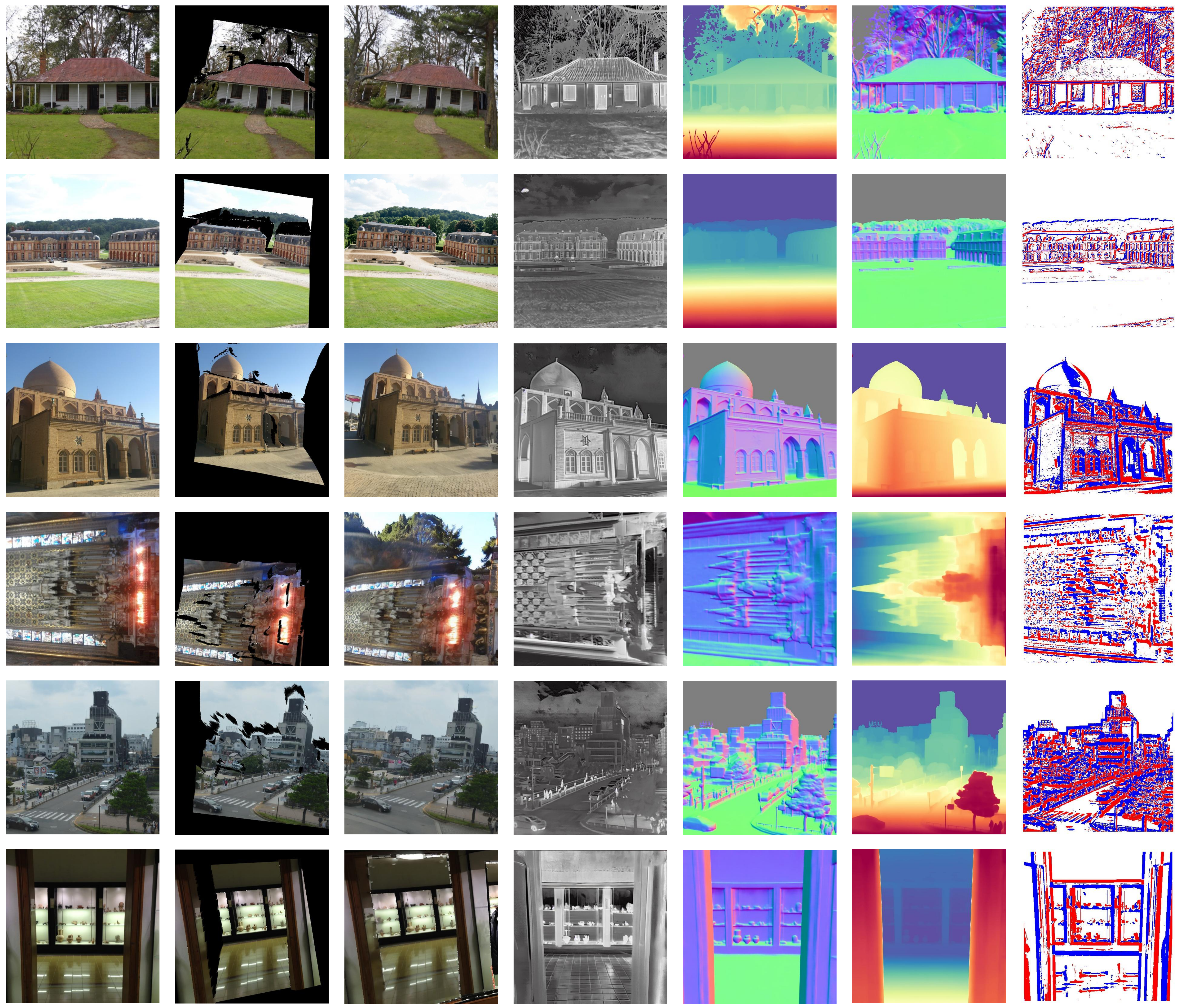}
	\caption{The first part of data synthesis results from Any-syn dataset. From left to right are the input image, novel-view image, inpainted image, pseudo-infrared image, pseudo-normal image, pseudo-depth image, and pseudo-event image.}\label{supp:fig1}
\end{figure*}

\begin{figure*}[t]	
	\centering	
	\includegraphics[scale=0.144]{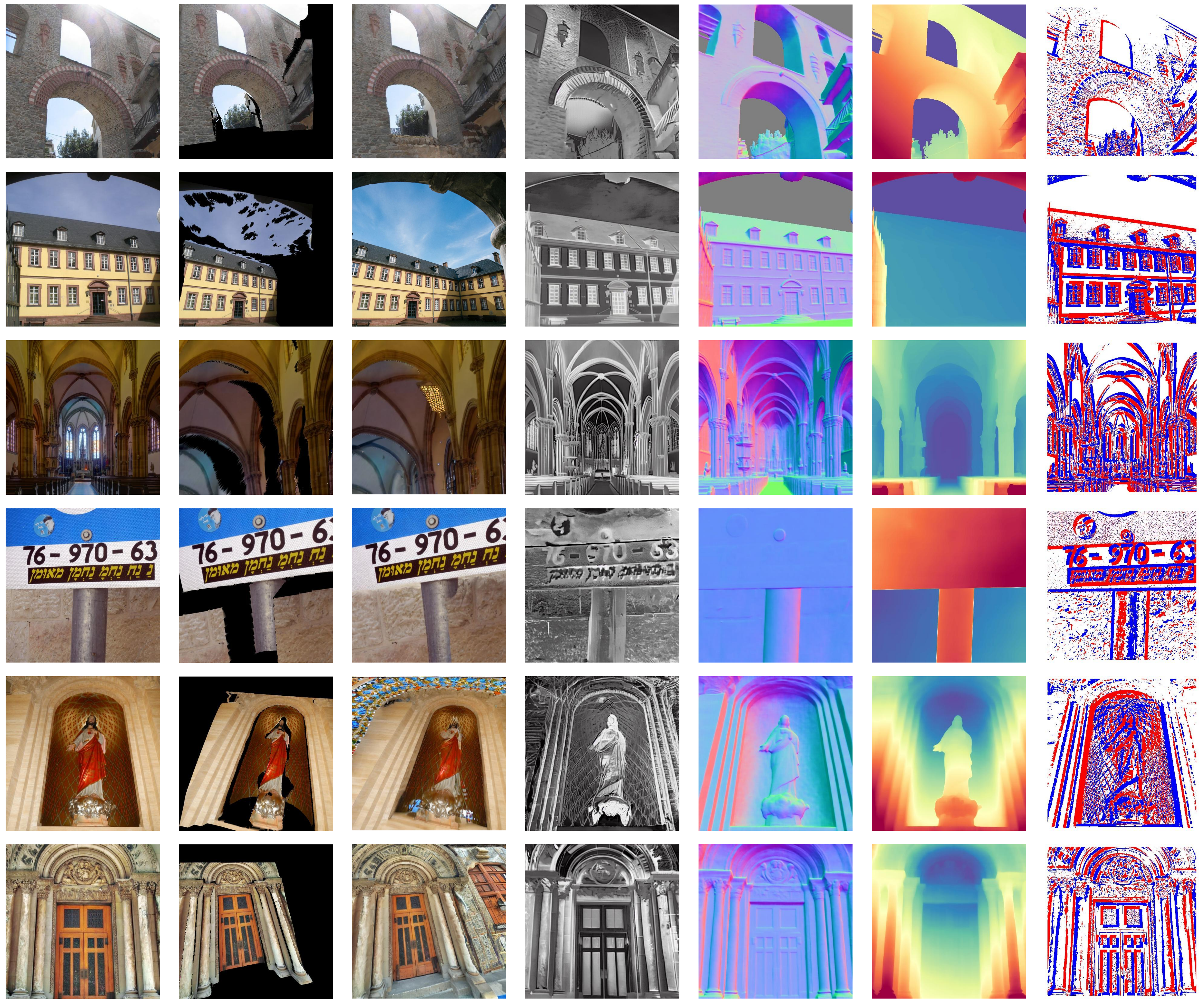}
	\caption{The second part of data synthesis results from Any-syn dataset. From left to right are the input image, novel-view image, inpainted image, pseudo-infrared image, pseudo-normal image, pseudo-depth image, and pseudo-event image.}\label{supp:fig2}
\end{figure*}

\begin{figure*}[t]	
	\centering	
	\includegraphics[scale=0.15]{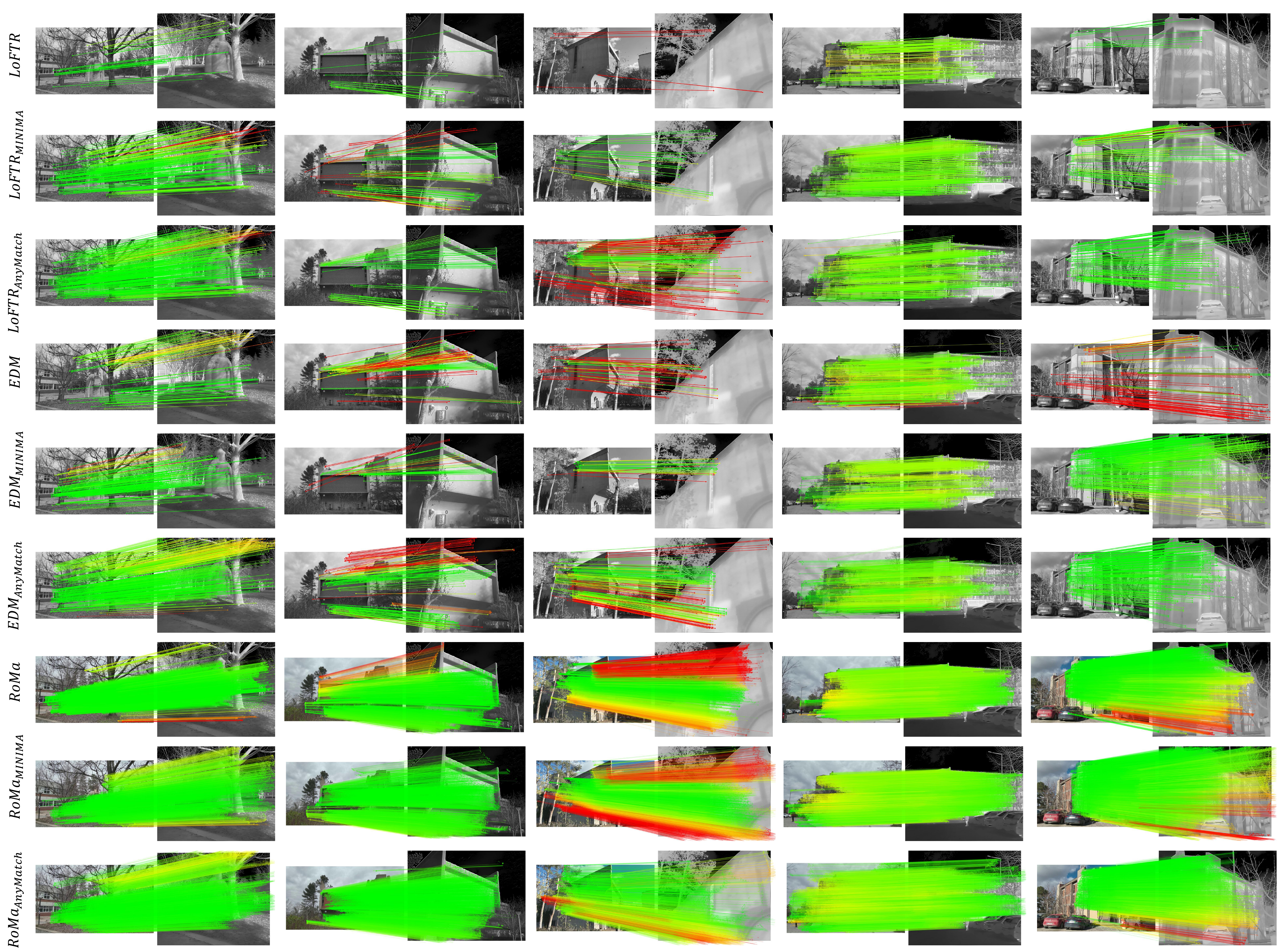}
	\caption{Qualitative results on real RGB-IR image pairs. Comparison of LoFTR, LoFTR$_\text{MINIMA}$, LoFTR$_\text{AnyMatch}$, EDM, EDM$_\text{MINIMA}$, EDM$_\text{AnyMatch}$, RoMa, RoMa$_\text{MINIMA}$, and RoMa$_\text{AnyMatch}$ across multiple cross-modal image pairs. Matches generated by each method are drawn, where the red lines indicate epipolar error (pose) or projection error (homography) beyond $5 \times 10^{-4}$ or 3 pixels.}\label{supp:fig3}
\end{figure*}

\begin{figure*}[t]	
	\centering	
	\includegraphics[scale=0.142]{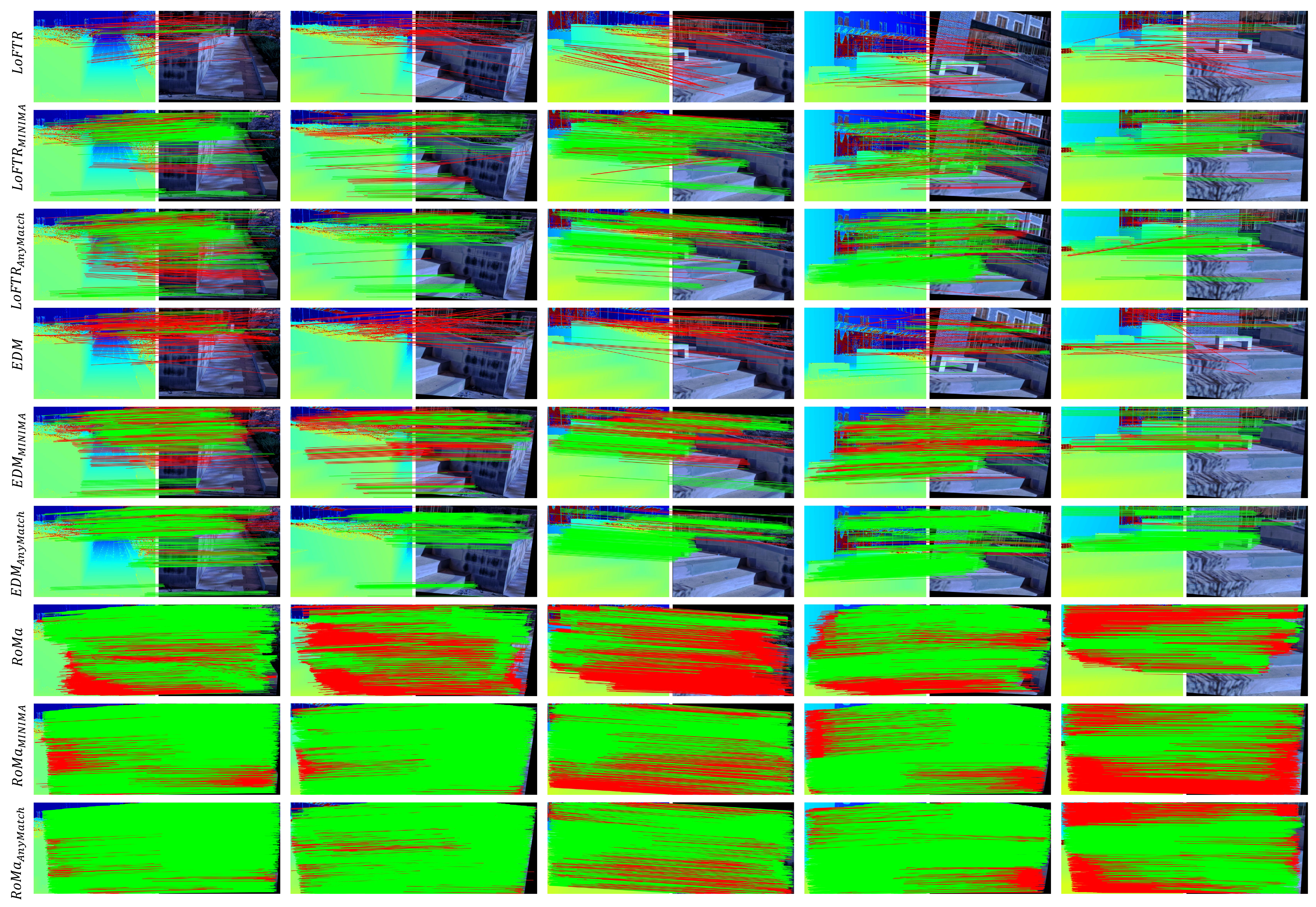}
	\caption{Qualitative results on real RGB-Depth image pairs. Comparison of LoFTR, LoFTR$_\text{MINIMA}$, LoFTR$_\text{AnyMatch}$, EDM, EDM$_\text{MINIMA}$, EDM$_\text{AnyMatch}$, RoMa, RoMa$_\text{MINIMA}$, and RoMa$_\text{AnyMatch}$ across multiple cross-modal image pairs. Matches generated by each method are drawn, where the red lines indicate epipolar error (pose) or projection error (homography) beyond $5 \times 10^{-4}$ or 3 pixels.}\label{supp:fig4}
\end{figure*}

\begin{figure*}[t]	
	\centering	
	\includegraphics[scale=0.142]{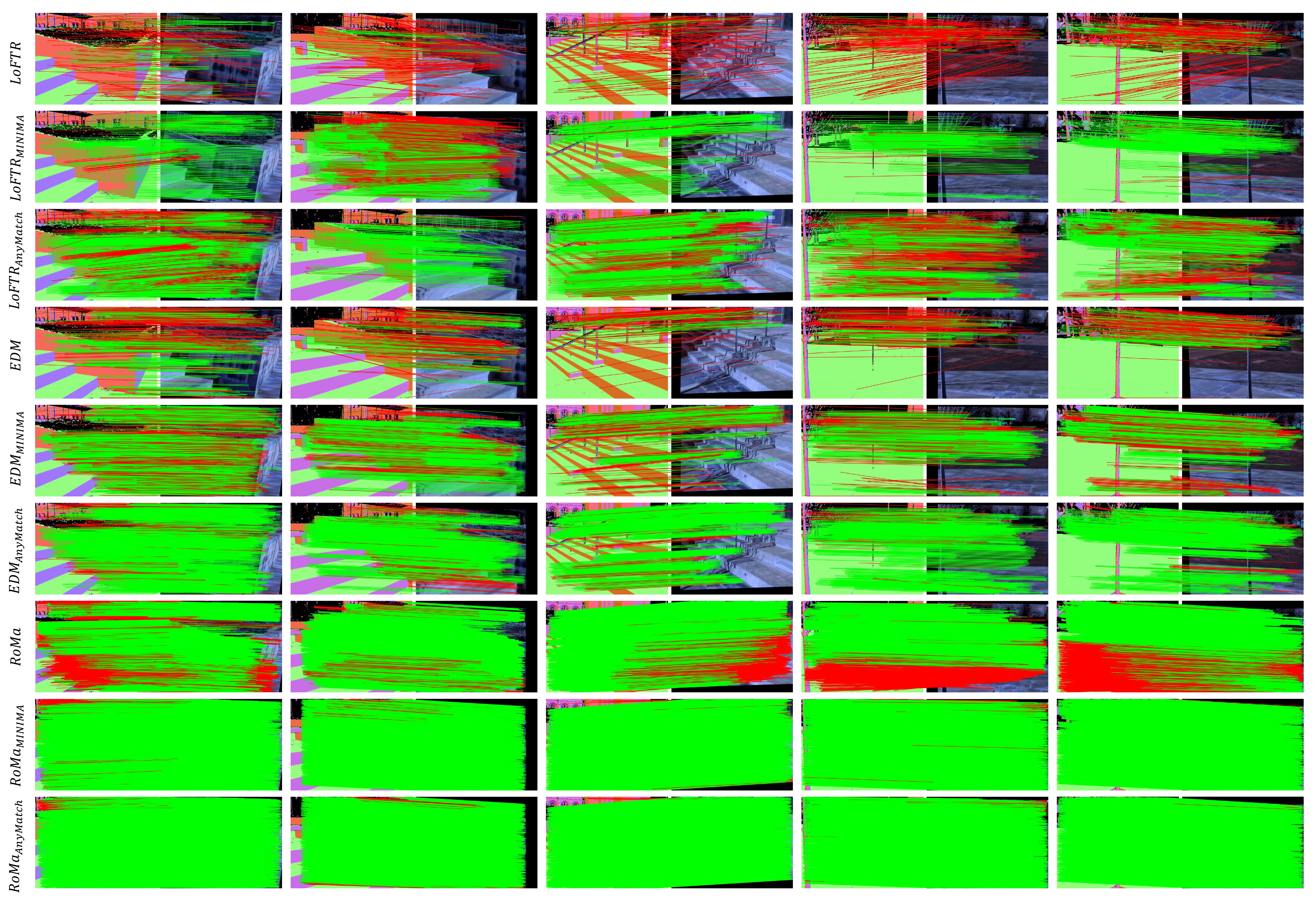}
	\caption{Qualitative results on real RGB-Normal image pairs. Comparison of LoFTR, LoFTR$_\text{MINIMA}$, LoFTR$_\text{AnyMatch}$, EDM, EDM$_\text{MINIMA}$, EDM$_\text{AnyMatch}$, RoMa, RoMa$_\text{MINIMA}$, and RoMa$_\text{AnyMatch}$ across multiple cross-modal image pairs. Matches generated by each method are drawn, where the red lines indicate epipolar error (pose) or projection error (homography) beyond $5 \times 10^{-4}$ or 3 pixels.}\label{supp:fig5}
\end{figure*}

\begin{figure*}[t]	
	\centering	
	\includegraphics[scale=0.142]{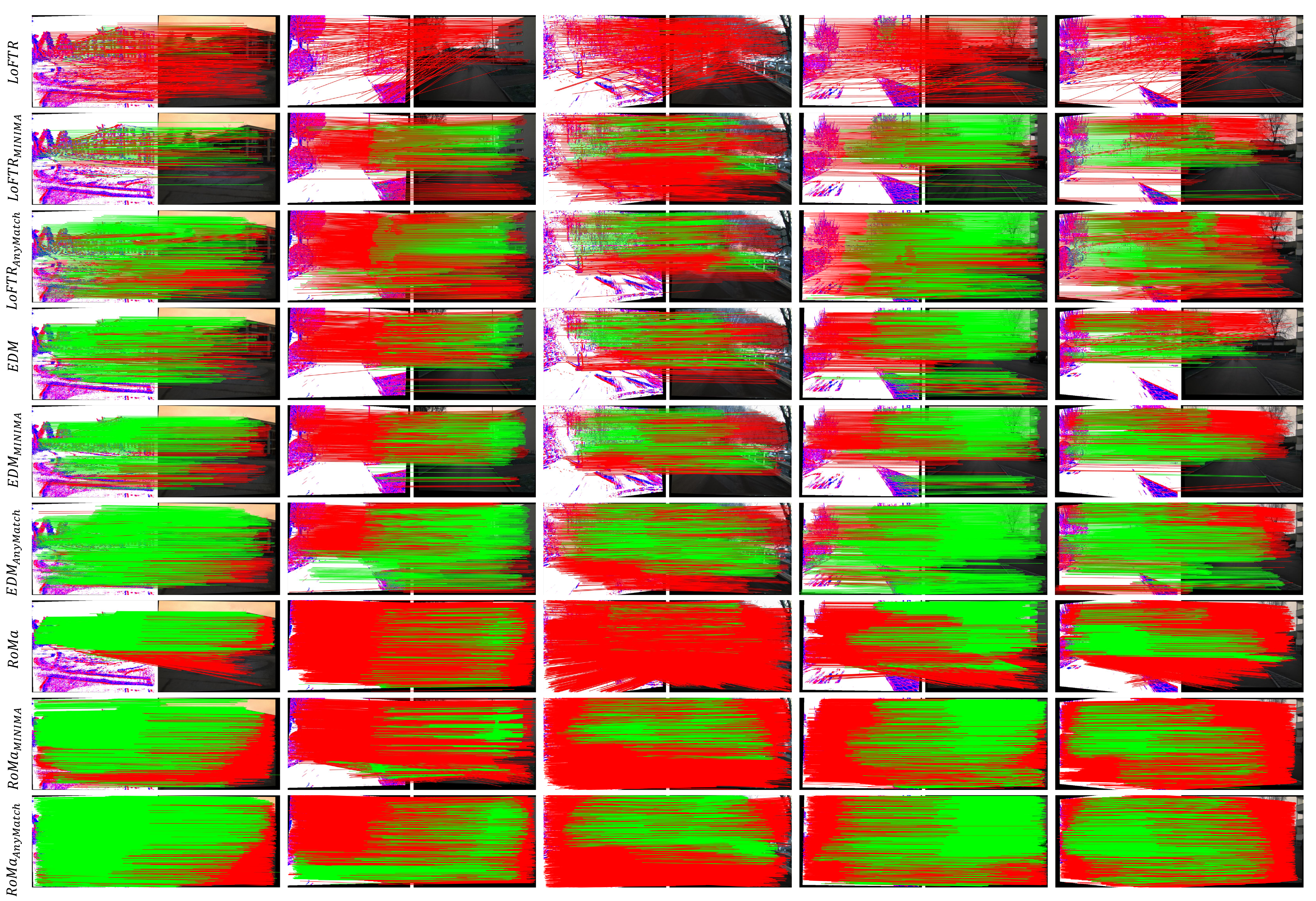}
	\caption{Qualitative results on real RGB-Event image pairs. Comparison of LoFTR, LoFTR$_\text{MINIMA}$, LoFTR$_\text{AnyMatch}$, EDM, EDM$_\text{MINIMA}$, EDM$_\text{AnyMatch}$, RoMa, RoMa$_\text{MINIMA}$, and RoMa$_\text{AnyMatch}$ across multiple cross-modal image pairs. Matches generated by each method are drawn, where the red lines indicate epipolar error (pose) or projection error (homography) beyond $5 \times 10^{-4}$ or 3 pixels.}\label{supp:fig6}
\end{figure*}

\begin{figure*}[t]	
	\centering	
	\includegraphics[scale=0.12]{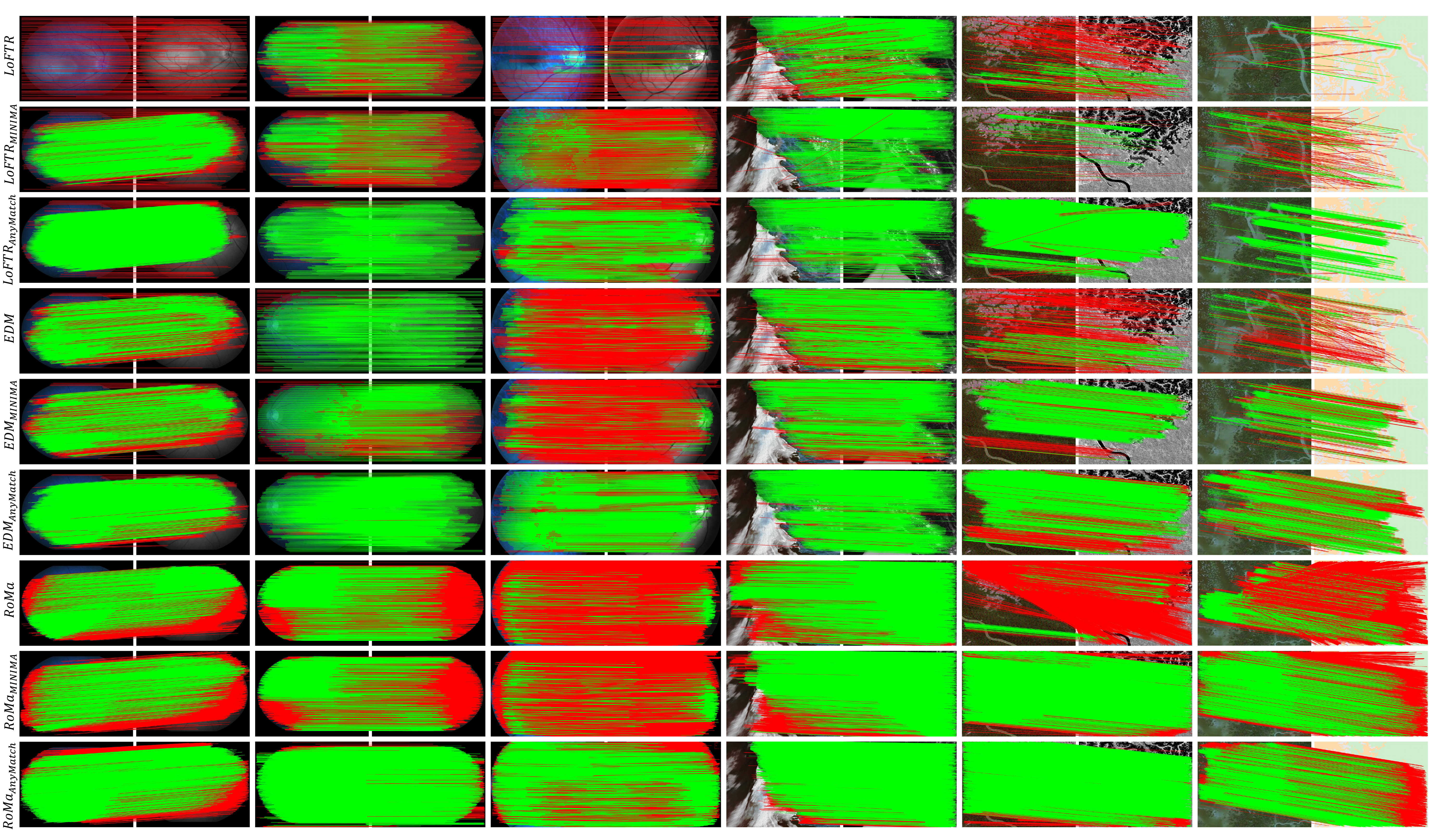}
	\caption{Qualitative results on real medical (retina) and remote sensing (including day-night, optical-infrared, and optical-map) image pairs. Comparison of LoFTR, LoFTR$_\text{MINIMA}$, LoFTR$_\text{AnyMatch}$, EDM, EDM$_\text{MINIMA}$, EDM$_\text{AnyMatch}$, RoMa, RoMa$_\text{MINIMA}$, and RoMa$_\text{AnyMatch}$ across multiple cross-modal image pairs. Matches generated by each method are drawn, where the red lines indicate epipolar error (pose) or projection error (homography) beyond $5 \times 10^{-4}$ or 3 pixels.}\label{supp:fig7}
\end{figure*}

In the SGCV module, we compute the PCK@5 for all synthesized image pairs in the Any-Syn dataset and analyze the distribution of PCK@5 across the entire dataset, as shown in Fig.~\ref{supp:fig12}. In Fig.~\ref{supp:fig13} and Fig.~\ref{supp:fig14}, we show representative synthesized image pairs with varying PCK@5. Fig.~\ref{supp:fig13} illustrates image pairs with relatively high PCK@5, which typically exhibit the following characteristics: (1) moderate geometric deformation between the image pair; (2) semantically plausible and realistic inpainted regions in the synthesized images, aligning well with semantic consistency; and (3) sufficient and effective matchable regions preserved in the synthesized images. In contrast, Fig.~\ref{supp:fig14} displays image pairs with lower PCK@5 scores, which are generally characterized by: (1) excessive geometric deformation between the image pair; (2) hallucination in the inpainted regions of synthesized images, violating semantic consistency; (3) inaccurate depth estimation leading to geometric inconsistency in the synthesized images; and (4) insufficient matchable regions available in the synthesized images.



\end{document}